\documentclass[11pt]{article}
\usepackage[]{acl}
\usepackage{times}
\usepackage{latexsym}
\usepackage{graphicx}
\usepackage{booktabs}
\usepackage{amsmath}
\usepackage{multirow}
\usepackage{subcaption}
\usepackage{enumitem}
\usepackage{algorithm}
\usepackage{algpseudocode}
\usepackage{amssymb}
\usepackage{pifont}
\usepackage[most]{tcolorbox}

\newtcolorbox{promptbox}[1]{
  colback=gray!5,
  colframe=gray!50,
  boxrule=0.5pt,
  arc=2pt,
  left=6pt,right=6pt,top=6pt,bottom=6pt,
  title={\small\bfseries #1},
  fonttitle=\bfseries,
  before skip=8pt,
  after skip=8pt,
  breakable
}

\title{Probing the Prompt KV Cache:
Where It Becomes Dispensable}

\author{
  \textbf{Vinayshekhar Bannihatti Kumar} \quad
  \textbf{Manoj Ghuhan Arivazhagan}\thanks{Equal contribution.} \quad
  \textbf{Disha Makhija}\footnotemark[\value{footnote}] \\
  \textbf{Rashmi Gangadharaiah} \\
  AWS AI Labs\\
  \texttt{\{vinayshk,mghuhan,dismakhi,rgangad\}}@amazon.com\\
}

\begin{document}

\maketitle

\newpage
\begin{abstract}
    Prior KV cache compression schemes empirically demonstrate that the
    prompt cache can be dropped or summarised during decoding with little accuracy loss, 
    indicating that it is partly redundant. We ask \emph{when} and
    \emph{what kind} of redundancy, i.e., at which layers, after how many decoding
    steps, and in what form can the prompt span KV cache be replaced
    without breaking the task. A controlled splice intervention swept over
    layers and decoding steps shows this redundancy is about
    \emph{form} rather than content. Replacing
    the upper layer prompt span KV cache with KV cache from a chat template
    scaffold with empty user content or filler tokens recovers near clean
    accuracy, while zeroing the entire prompt KV collapses accuracy. The
    dissociation replicates across the Qwen3, Gemma 3, and Llama 3 model families of varying sizes
    on multiple datasets. A serving implementation of this technique reduces KV memory by $15.4\%$, supports $18\%$ more HBM-constrained concurrent requests, and improves throughput by
$13$--$21\%$. 
\end{abstract}

\section{Introduction}
\label{sec:intro}

A growing body of work hints that LLMs use the prompt differently at different stages of the forward pass. Attention concentrates on
the first few positions as a positional
sink~\citep{xiao2024streamingllm}, the same massive activations
explain mid-layer compression valleys~\citep{sun2025compressionvalleys},
in-context demonstrations are summarised into compact internal task
and function
vectors~\citep{hendel2023taskvectors,todd2024functionvectors}.
Cross-prompt KV substitution works show that generation can be
hijacked by overwriting the trailing token positions across all layers
of the cache with another prompt's
KV~\citep{ganesh2025historyswapping}, though that work studies attack
feasibility rather than what the cache encodes. KV-cache
compression
schemes~\citep{cai2024pyramidkv,li2024snapkv,zhang2023h2o,ge2024fastgen,liu2024scissorhands}
go further and empirically show that large fractions of the prompt cache
can be dropped or summarised with little accuracy loss, indicating
partial redundancy. Each line of work, however, operates
at a narrow scope (four sink tokens, BOS-token activations, a single
task vector, an all-layer-or-nothing swap with one donor pair, or a
specific pruning policy). It leaves open when and with what the prompt-span KV cache can be replaced, namely at which layers, after how many decoding steps, and with what donor content. The answer determines if compression schemes must preserve, drop, or substitute these entries.




We address this with a controlled splice intervention swept over layer cutoff and decoding steps with donor caches varying in how much of the prompt's form and content (the user's task) they preserve,
  tracing a phase diagram of when the prompt-span KV cache can be replaced without breaking the task. We define \emph{form} as the chat-template scaffolding required to structure
  a conversation, namely the special tokens that delimit roles and turns, such
  as \texttt{<|im\_start|>}, \texttt{system}, \texttt{user},
  \texttt{assistant}, etc.
  We study how the prompt KV cache shapes generation by mapping where and in what way the cache is dispensable, we provide insights intended to inform the design of compression methods.


Our contributions are: (i) The first systematic characterisation of
\emph{when} the decode-time prompt KV cache becomes dispensable. (ii)
\emph{What} of the prompt the cache must retain, via a
form-vs-content dissociation showing that chat-template scaffolding is
sufficient and task content is dispensable. (iii) A donor-noise ladder built on a 180-variant algorithmic-donor
  benchmark we designed to vary donor quality systematically. Moving along the
  ladder (\textsc{same-algo} $\to$ \textsc{diff-algo} $\to$
  \textsc{diff-family}) injects progressive noise into the donor cache,
  letting us measure how performance degrades as the donor content drifts
  further from the target task. Together, these results provide an end-to-end picture: at upper layers,
form alone is sufficient, whereas at lower layers the cache content is
causally consulted, with recovery cost increasing as the donor drifts from
the target task. These findings also translate into practical serving gains,
reducing KV memory by $15.4\%$, supporting $18\%$ more concurrent requests
under HBM constraints, and improving throughput by $13$--$21\%$, with only
a $1$--$2$ point performance drop on LIFBench across 4k, 8k, and 16k-token
contexts.

\begin{figure*}[t]
\centering
\begin{subfigure}[t]{0.49\linewidth}
\centering 
\includegraphics[width=\linewidth]{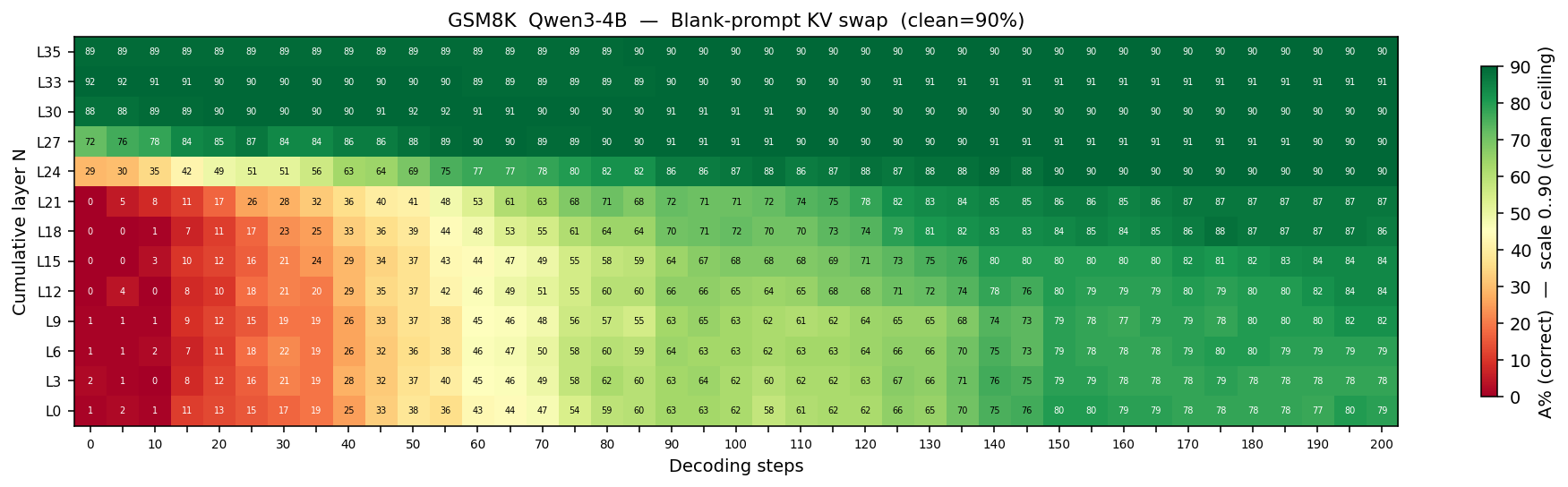}
\caption{GSM8K \textsc{BLANK}}
\end{subfigure}\hfill
\begin{subfigure}[t]{0.49\linewidth}
\centering
\includegraphics[width=\linewidth]{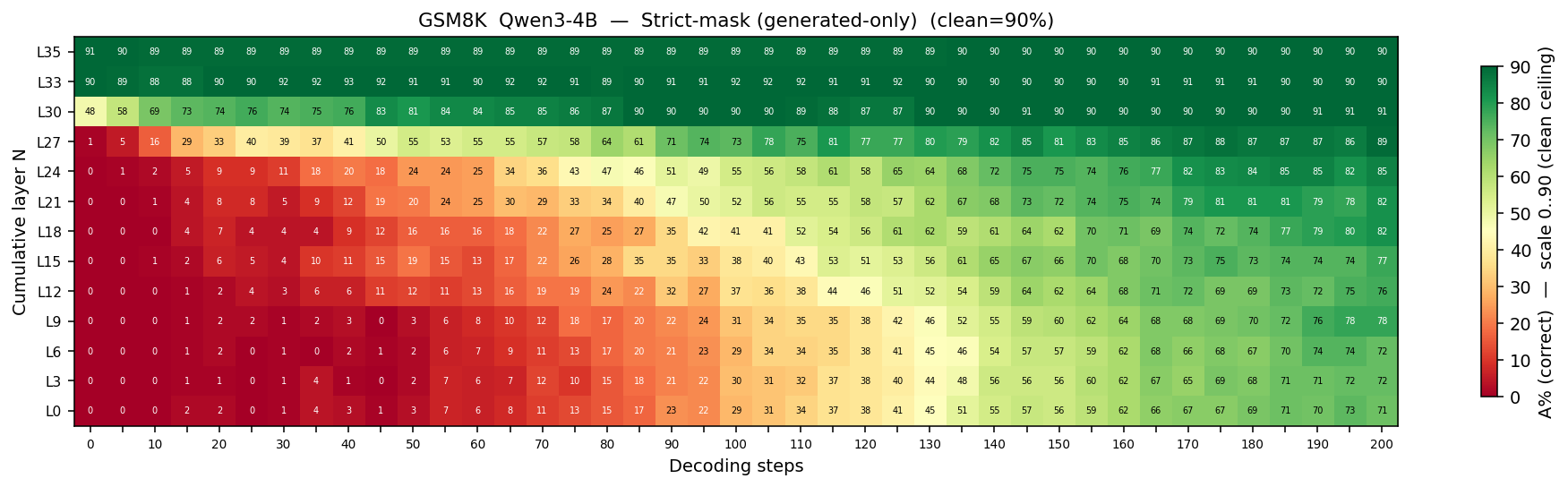}
\caption{GSM8K \textsc{ZERO}}
\end{subfigure}\\[0.4em]
\begin{subfigure}[t]{0.49\linewidth}
\centering
\includegraphics[width=\linewidth]{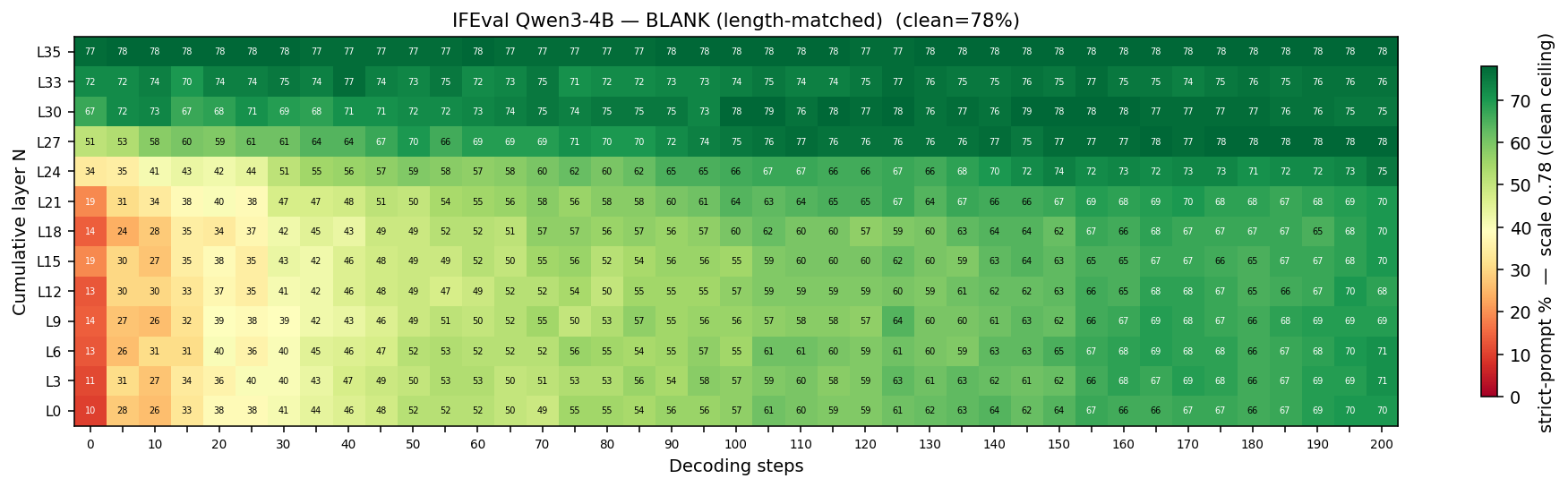}
\caption{IFEval \textsc{BLANK}}
\end{subfigure}\hfill
\begin{subfigure}[t]{0.49\linewidth}
\centering
\includegraphics[width=\linewidth]{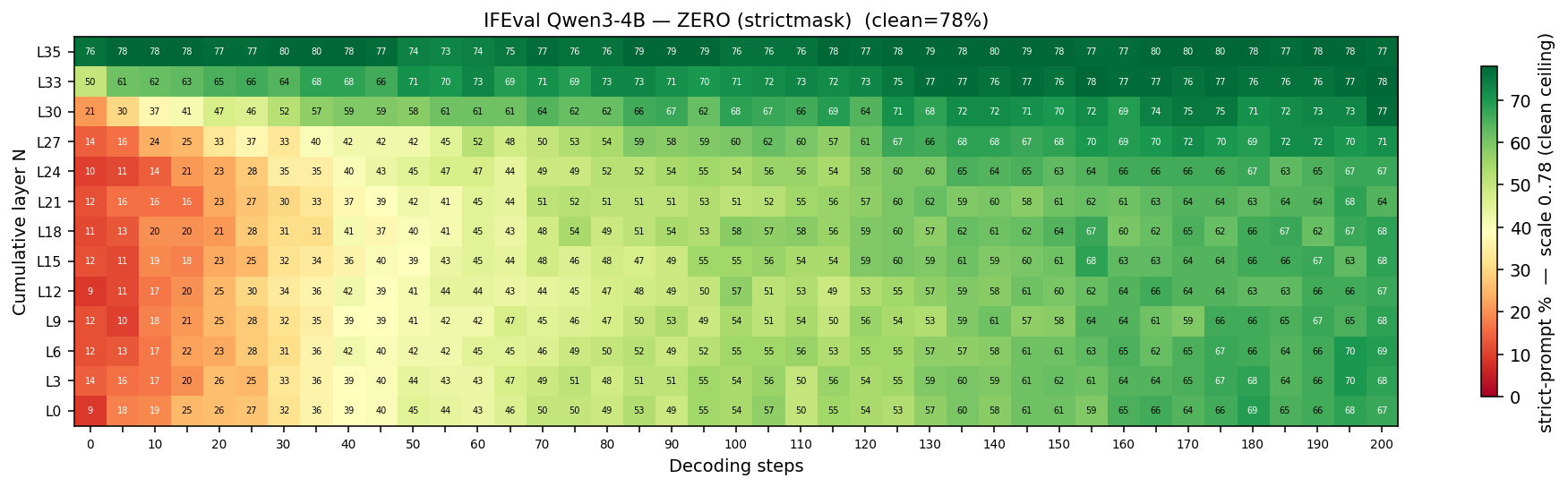}
\caption{IFEval \textsc{ZERO}}
\end{subfigure}
\caption{Qwen3-4B heatmaps for \textsc{zero} and \textsc{blank}
on GSM8K (top) and IFEval (bottom). Each cell shows pass\% at one
$(L, W)$. \textsc{Blank} recovers far faster than \textsc{zero} along
both $L$ and $W$.}
\label{fig:phase-diagrams-qwen3-4b}
\end{figure*}

\section{Models and Datasets}
\label{sec:datasets}
We evaluate Qwen3-4B, Qwen3-8B, Qwen3-32B, Gemma-3-4B-IT, and
Llama-3-8B-Instruct~\citep{qwen3,gemma3,grattafiori2024llama3} with greedy
decoding on four datasets: \textbf{GSM8K}~\citep{cobbe2021gsm8k}
(chain-of-thought arithmetic), \textbf{MBPP}~\citep{austin2021mbpp} and
\textbf{HumanEval}~\citep{chen2021humaneval} (Python code completion), and
\textbf{IFEval}~\citep{zhou2023ifeval} (open-ended generation). We sample 100
problems from each. Scoring details are in Appendix~\ref{app:datasets}.
We additionally evaluate \textbf{LIFBench}~\citep{lifbench}
(long-context instruction following) across all eleven tasks at approximately
4k, 8k, and 16k tokens. We sample 150 examples per context length and use
the benchmark's official judges.

\textbf{Algorithmic-donor benchmark.} A Python benchmark of 9 algorithmic
families (search, sort, graph traversal, shortest path, fibonacci,
max-subarray, primes, two-sum, string-match), 10 problems each. Every problem
is posed in two variants that share the underlying task but pin the solution
to a different algorithm class (e.g.\ linear- vs.\ binary-search), giving 180
variants indexed by a (problem $p$, algorithm-class $c$, family $f$) triple,
each with a hidden test suite. We score correctness as test-suite pass plus a
target-algorithm match check by LLM judge. The details about the dataset construction and a human
validation of the benchmark and judge are in Appendix~\ref{app:algo-bench}.

\section{Experimental Setup}
\label{sec:setup}

\paragraph{Splice intervention.}
We study chat-tuned decoder transformers with $N$ layers. Prefilling a
prompt of length $T_p$ yields a KV cache
$K^{(\ell)}, V^{(\ell)} \in \mathbb{R}^{T_p \times d}$ at each layer $\ell$,
which the model attends over (together with generated-span entries) during
decoding.

Every experiment is a single intervention with three parameters: a layer
cutoff $L$, a splice onset $W$, and a donor cache $\tilde{K}, \tilde{V}$. We
patch the layer set $\mathcal{P} = \{\ell : \ell \geq L\}$. The initial token is generated as part of the prefill. For decoding steps
$t \leq W$ the unmodified cache is used, and at step $t = W + 1$ the prompt-span
entries at every $\ell \in \mathcal{P}$ are replaced in place
($K^{(\ell)}_{0:T_p} \leftarrow \tilde{K}^{(\ell)}_{0:T_p}$, likewise for
$V$), after which decoding continues to completion. Generated-span entries are
never modified. The donor takes one of seven forms
(Table~\ref{tab:donors}); for prompt-derived donors we length-match target and
donor at the token level so the patched slots stay positionally aligned. The \textsc{scaffold-only} and $\textsc{blank}_{\rho}$ are $k$-token donors ($k{=}17$ for
  \textsc{scaffold-only} on Qwen3) prefilled at positions $[T_p{-}k,\,T_p)$
  and written into the final $k$ prompt-span slots.

\begin{table}[t]
\small
\centering
\setlength{\tabcolsep}{4pt}
\begin{tabular}{@{}l p{0.66\linewidth}@{}}
\toprule
donor & cache content \\
\midrule
\textsc{zero}        & all-zero $K, V$ at every patched layer \\
\textsc{blank}       & chat-template scaffolding with user content replaced by filler token ``a'' to length $T_p$ \\
\textsc{scaffold only} & \textsc{blank} with the filler removed (scaffold only) \\
\textsc{blank}$_{\rho=0.5}$ & \textsc{blank} with filler reduced to a fraction $\rho$ of the span ($\rho{=}0.5$) \\
\textsc{same-algo}   & donor variant $(p', c, f)$ \\
\textsc{diff-algo}   & donor variant $(p', \neg c, f)$ \\
\textsc{diff-family} & donor variant $(\cdot, \cdot, f')$ \\
\bottomrule
\end{tabular}
\caption{Donor caches. All interventions modify the same prompt-span
positions at the same layers under the same splice onset; only the
content varies. Algorithmic-donor variants are indexed by a triple
$(p, c, f)$ of problem $p$, algorithm-class $c$, family $f$.}
\label{tab:donors}
\end{table}

\paragraph{Sweep and reporting.}
For every (model, task, intervention) we run the full grid
$L \in \{0, 3, 6, \ldots, N{-}1\}$, $W \in \{0, 5, 10, \ldots, 200\}$
plus the unintervened \emph{clean}
baseline. We summarise each condition as (i) an $L \times W$ phase-diagram
heatmap and (ii) a \emph{recovery frontier}
$W^\star(L; \alpha) = \min \{W : \mathrm{pass}(L, W) \geq \alpha\,p_{\mathrm{clean}}\}$,
where $p_{\mathrm{clean}}$ is the unintervened pass rate of the same model
on the same task and $\alpha \in (0, 1]$.
\section{Results and Analysis}
\label{sec:results}
\label{sec:analysis}

\begin{figure}[t]
\centering
\includegraphics[width=\linewidth]{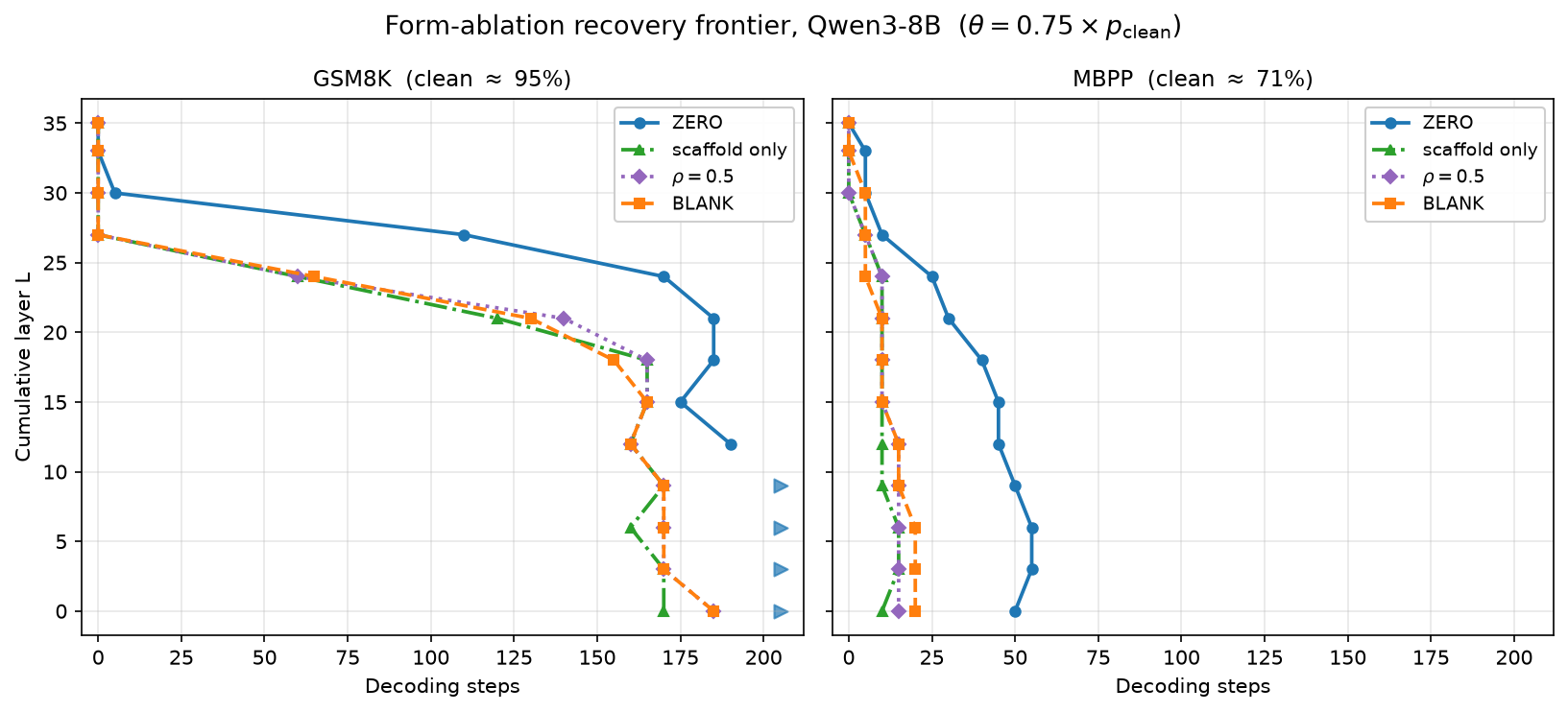}
\caption{Form-component ablation on Qwen3-8B: recovery frontier
$W^\star(L; \alpha\!=\!0.75)$ on GSM8K and MBPP. The role
markers are held fixed while the filler between them varies
(\emph{scaffold only}, $\rho{=}0.5$, \textsc{blank}). All three coincide and
sit far left of \textsc{zero}: the role-marker scaffold, not its length
or filler, drives recovery.}
\label{fig:form-ablation-frontier}
\end{figure}

\begin{figure*}[t]
\centering
\includegraphics[width=\linewidth]{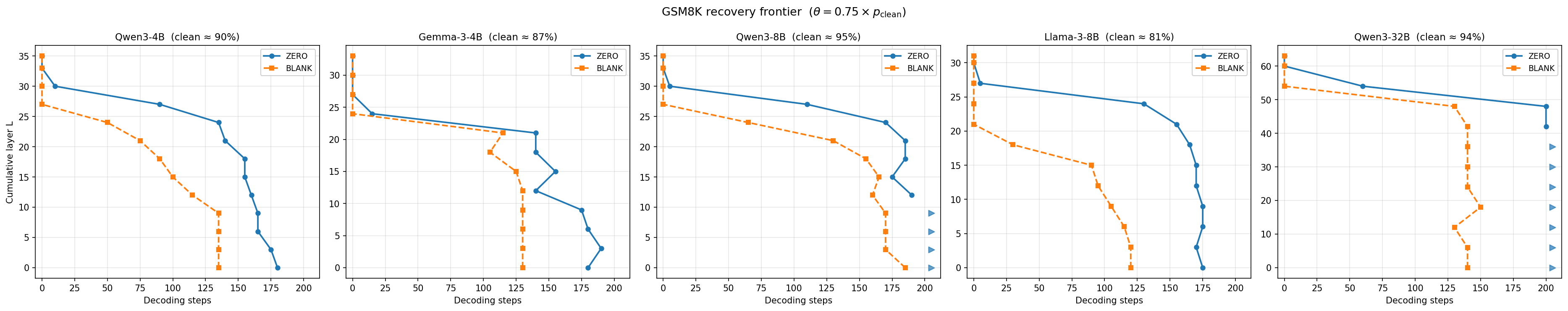}
\caption{GSM8K recovery frontier $W^\star(L; \alpha\!=\!0.75)$ for
\textsc{zero} vs \textsc{blank}, one panel per model
(Qwen3-4B to Qwen3-32B). \textsc{Blank} sits consistently left of
\textsc{zero}; the gap widens with scale, and on Qwen3-32B \textsc{zero}
never recovers below $L\!\approx\!42$ (right-edge markers: no recovery
within 200 steps).}
\label{fig:strict-vs-blank-frontier}
\end{figure*}

\begin{figure*}[t]
\centering
\includegraphics[width=0.8\linewidth]{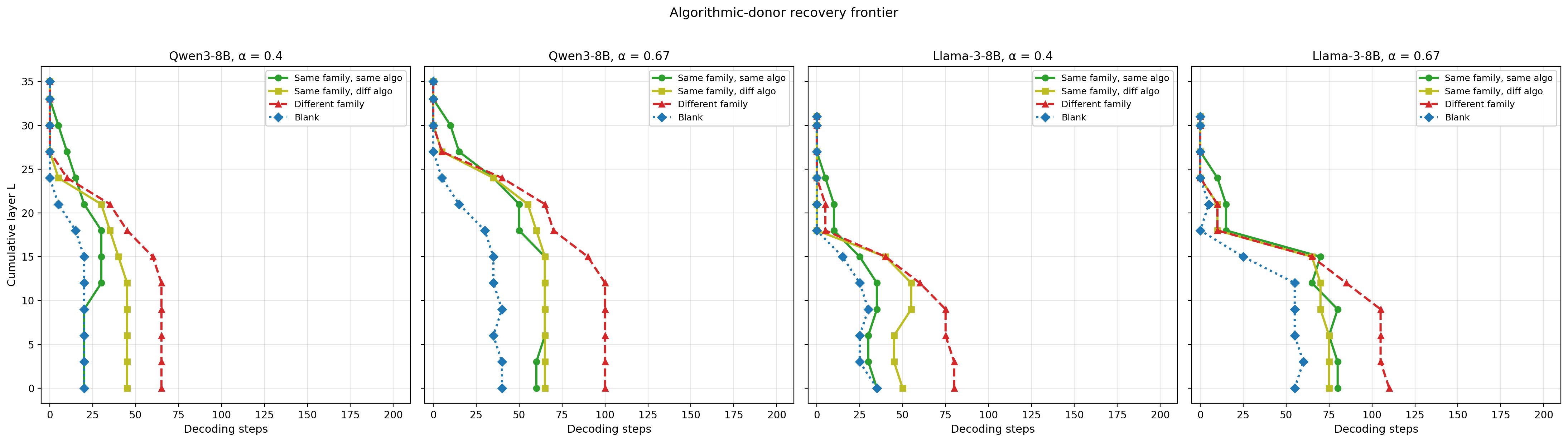}
\caption{Recovery frontier $W^\star(L; \alpha)$ on the algorithmic-donor
benchmark for Qwen3-8B and Llama-3-8B-Instruct at
$\alpha\!\in\!\{0.4, 0.67\}$. Recovery cost grows with donor
noise: \textsc{diff-family} is highest, \textsc{blank} lowest.}
\label{fig:noise-ladder-frontier}
\end{figure*}

\paragraph{(R1) Form is sufficient, content is dispensable.}
We probe whether the prompt cache content matters at upper layers in two
steps. We first ask whether \emph{task content} is required at all by
running the \textsc{blank} intervention (Table~\ref{tab:donors}).
Recovery is rapid on both axes (Fig.~\ref{fig:phase-diagrams-qwen3-4b}).
On Qwen3-4B GSM8K (clean $=90\%$, median length $=147$), \textsc{blank} reaches $88\%$ at
$L{=}30, W{=}0$. Sweeping along $W$ at
fixed $L{=}27$, it hits $87\%$ by $W{=}25$. IFEval (clean $=78\%$, median length $=261$), MBPP (clean $=71\%$, median length $=79$), and HumanEval (clean $=79\%$, median length $=121$) reproduce
the shape observed in GSM8K (Appendix~\ref{sec:appendix-heatmaps},~\ref{sec:appendix-mbpp-frontier},~\ref{sec:appendix-humaneval}).
This shows that retaining the prompt's form, while discarding its task content, is sufficient at upper layers.

If task content is not required, what is? We push further and remove the scaffolding too via \textsc{zero}. Recovery still happens,
but is markedly slower on both axes than \textsc{blank}. At Qwen3-4B GSM8K $W=0$, \textsc{zero} sits at $1\%$ at
$L{=}27$, climbs to $48\%$ at $L{=}30$, and only matches clean at $L{=}33$. Along $W$ at $L{=}27$, it crosses $87\%$ only near
$W{\approx}170$, seven times the pre-splice decoding steps compared to \textsc{blank}. IFEval, MBPP and HumanEval also show this gap. The two
interventions converge at the boundaries. As seen in Fig.~\ref{fig:phase-diagrams-qwen3-4b} both match clean at $L{=}33$ and both collapse at shallow
$L$ with $W{=}0$, where no substituted cache can compensate for missing
prompt context the model has not yet seen. 

Which part of the \textsc{blank} donor carries the load? \textsc{blank} keeps the
 scaffold and fills the rest of the prompt span with filler.
On Qwen3-8B we vary only how much filler sits between the markers,
holding the markers fixed. A bare scaffold with no filler and a \textsc{blank}
half-filled ($\rho{=}0.5$) donor each recovers as fast as \textsc{blank} at
every $L$ on GSM8K and MBPP, all far faster than \textsc{zero}
(Fig.~\ref{fig:form-ablation-frontier}) indicating  what the upper layers need is the
chat-template scaffold itself. Neither the filler content nor the filled length matters. What the upper layers need is the role-marker scaffold itself.

\paragraph{(R2) Phase diagram across models and tasks.}
The (R1) shape generalises beyond a single model and task. The recovery
frontier $W^\star(L; \alpha\!=\!0.75)$ in
Fig.~\ref{fig:strict-vs-blank-frontier} plots \textsc{zero}  and
\textsc{blank} on GSM8K across five models spanning Qwen3-4B to
Qwen3-32B. \textsc{Blank} recovers faster, sitting consistently to the
left of \textsc{zero} in every panel. At intermediate $L$ the gap is
large, on Qwen3-4B GSM8K,
$\sim$60 steps at $L=15$ widening to $\sim$90 steps at $L=24$, and
both lines collapse to $W=0$ only at the topmost cutoffs (\textsc{blank}
from $L \approx N{-}6$ onwards, \textsc{zero} a few layers deeper). The
dissociation persists and sharpens with scale: on Qwen3-32B (64 layers)
\textsc{blank} recovers at every cutoff while \textsc{zero} never reaches
threshold below $L\!\approx\!42$ within 200 steps. The same pattern holds
on MBPP (App.~\ref{sec:appendix-mbpp-frontier}) and on the
algorithmic-donor benchmark, where the \textsc{blank} intervention
recovers faster than any other donor type
(Fig.~\ref{fig:noise-ladder-frontier}).

\paragraph{(R3) Donor noise determines lower-layer recovery cost.}
At lower layers the prompt cache is no longer dispensable: what gets
injected matters, and how much it matters tracks how far the donor cache
is from the target. Fig.~\ref{fig:noise-ladder-frontier} plots the
recovery frontier $W^\star(L; \alpha)$ on the algorithmic-donor
benchmark for Qwen3-8B and Llama-3-8B-Instruct at two thresholds
($\alpha = 0.4, 0.67$). 
\textsc{Blank} is closest to the $W{=}0$ axis at every $L$,
consistent with (R1): when no task content is injected, decoding
recovers fastest. The donor curves degrade as the donor's content
drifts from the target task. The within-family donors (\textsc{same-algo}
and \textsc{diff-algo}) track each other and require modest pre-splice
decoding to recover even at shallow $L$; the across-family donor
(\textsc{diff-family}) sits consistently above them, requiring
substantially more $W$ to clear the same threshold. The pattern holds
for both Qwen3-8B and Llama-3-8B-Instruct and is sharper at the
stricter $\alpha=0.67$, where the gap between \textsc{diff-family} and
the within-family lines widens to tens of decoding steps. Together with
(R1), these results give an end-to-end picture: at upper layers the
model needs only the prompt's structural form, while at lower layers it
actively processes the cache content, with quality degrading as the
injected content moves further from the target task. 
See the Appendix for per-model phase diagrams~\ref{sec:appendix-heatmaps} and algorithmic-donor heatmaps~\ref{sec:appendix-algo-heatmaps}, splice pseudocode~\ref{sec:appendix-pseudocode}, and example generations~\ref{sec:appendix-examples}.

\paragraph{(R4) The result extends to long contexts.}
The \textsc{scaffold-only} intervention also remains effective on LIFBench.
With $L{=}30$ and $W{=}20$, performance differs from unmodified decoding by
$0.9$, $2.0$, and $1.4$ points at 4k, 8k, and 16k tokens, respectively
(Fig.~\ref{fig:serving}, right). The gap does not increase with context length, showing that upper-layer prompt KV remains largely
dispensable even for long-context instruction-following tasks.

\section{Efficient Serving with Shared Scaffold KV}
\label{sec:serving}

The results in Section~\ref{sec:results} suggest a concrete serving strategy.
At lower layers $\ell < L$, we retain the request's prompt KV throughout
decoding because its task content remains important. At upper layers
$\ell \ge L$, we retain the complete prompt KV during the warmup steps
$t \leq W$ and replace it with a content-free scaffold at step $t=W + 1$. This
section evaluates whether this depth- and time-dependent intervention can
produce practical serving gains.

\paragraph{Serving design.}
We implement this policy in a paged-KV serving prototype. Prefill and the
first $W$ decoding steps proceed normally. At step $W$, the page tables for
layers $\ell \ge L$ are changed from
\texttt{[prompt $|$ generated]} to
\texttt{[scaffold $|$ generated]}. The request's original prompt pages at
these layers can then be freed. Prompt KV at layers $\ell < L$ and all KV
for generated tokens remain request-specific and are never modified. The scaffold contains 17 tokens (on Qwen3 models) and no task-specific content.
  Consequently, requests of the same prompt length can point to the same
  scaffold pages instead of storing separate copies. We construct one scaffold
  cache per prompt length and place its tokens at the final positions of the
   prompt span so that their rotary positions align. Because rotary embeddings encode absolute position, each prompt length $T_p$ requires a separate scaffold cache. 
  
\paragraph{Memory and latency.}
We evaluate the configuration $L{=}30$, $W{=}20$ on Qwen3-8B. After the
splice, eliminating the upper-layer prompt pages reduces per-request KV
memory by $15.4\%$ for an 8k-token prompt. When concurrency is constrained by HBM capacity, this reduction allows
$18\%$ more requests to run concurrently under the same KV-memory budget.

The splice also shortens the upper-layer attention span. Instead of
attending to $T_p{+}t$ cached tokens, these layers attend to only
$17{+}t$, where $T_p$ is the prompt length and $t$ is the number of
generated tokens. At a matched batch size, this reduces kernel-bound
decoding latency by $7\%$ on an A100-40GB GPU with CUDA graph replay.

\begin{figure}[t]
  \centering
  \includegraphics[width=0.5\textwidth]{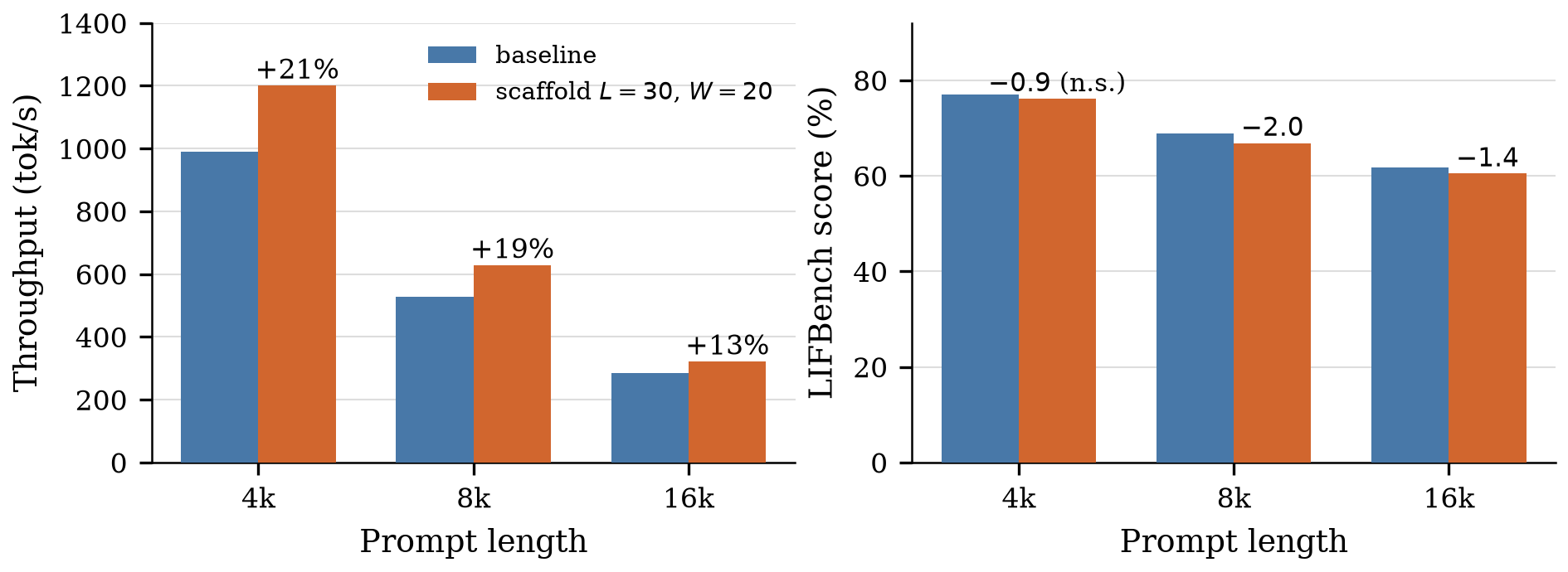}
  \caption{Serving results for scaffold KV sharing with $L{=}30$ and
  $W{=}20$ on Qwen3-8B.
  \textbf{Left:} Decode throughput at each method's memory-limited batch
  size. Sharing the upper-layer scaffold reduces per-request KV memory,
  permits larger batches, and shortens upper-layer attention.
  \textbf{Right:} LIFBench accuracy shows no meaningful regression.}
  \label{fig:serving}
\end{figure}

These effects improve throughput in complementary ways. Lower memory usage
permits a larger batch under the same KV budget, while the shorter attention
span reduces the cost of decoding that batch. At the memory-limited
operating point, the resulting throughput improvement is $13$--$21\%$
tokens per second across the evaluated context lengths
(Fig.~\ref{fig:serving}, left).

\section{Related Work}

Prior work documents prompt-cache redundancy at narrow scopes,
including positional sinks at the first $\sim$4
tokens~\citep{xiao2024streamingllm}, mid-layer compression valleys
driven by massive activations~\citep{sun2025compressionvalleys}, the
lost-in-the-middle effect in which models underperform on relevant
content placed in the middle of long
contexts~\citep{liu2024lostmiddle}, and ICL summarisation into compact
internal task and function
vectors~\citep{hendel2023taskvectors,todd2024functionvectors}. A
complementary mechanistic line argues that in-context copying is
implemented by induction heads, a distributed attention pattern rather
than a vector summary~\citep{olsson2022inductionheads}. KV-cache
compression schemes~\citep{cai2024pyramidkv,li2024snapkv,zhang2023h2o,ge2024fastgen,liu2024scissorhands}
prune by attention magnitude, and our results suggest a complementary
axis of substitution by structural placeholders at upper layers.
\citet{ganesh2025historyswapping} is the closest methodologically, an
attack feasibility study that hijacks generation by overwriting
trailing token positions across all KV layers with fixed donor. We
ask a different question and run a graded donor-noise ladder, expose
an algorithmic-overlap inversion, and add the form-vs-content
\textsc{zero}-vs-\textsc{blank} dissociation.

\section{Conclusion}
A controlled splice intervention across layers and decoding steps
reveals a depth-dependent split in what the prompt KV cache must carry.
At upper layers, chat-template alone is sufficient:
content-free filler recovers near-clean accuracy while zeroing the same
slots collapses generation. At lower layers, content is causally
active: donor caches degrade accuracy in proportion to their distance
from the target task, with cross-family donors requiring substantially
more decoding steps to recover than within-family donors. For
KV-cache compression, the practical implication is that upper-layer
prompt entries can be replaced with structural placeholders rather than
dropped, while lower-layer entries must preserve task-specific content.
\section*{Limitations}

Our intervention modifies only the prompt-span KV cache, which remains static
during decoding; generated-token KV entries are never altered. Although the
mechanistic results replicate across multiple model families, our serving
evaluation is limited to Qwen3-8B on an A100-40GB GPU using a paged-KV
prototype. The exact memory, latency,
concurrency, and throughput gains may therefore vary across models, hardware,
serving backends, and workloads. In particular, the gains depend on prompt
and generation lengths and are most useful when concurrency is constrained
by HBM capacity. We also restrict our study to
chat-tuned decoder-only transformers and do not claim that the findings
transfer to base models, encoder-decoder architectures, or substantially
larger models. Future work should evaluate these regimes and validate the
serving gains across additional model families and accelerator platforms.

\section*{Use of LLM in writing this paper}
Large language models were used as productivity tools during the preparation of this work. Specifically, we used LLM-based assistants to: (1) generate LaTeX tables from CSV files, with all numerical values manually verified against the source data; (2) assist with grammar, formatting, and rephrasing of text to meet page limits; (3) generate initial code for experimental scripts, evaluation pipelines, and data processing utilities; (4) draft and iterate on LaTeX figure and table layouts; and (5) assist with debugging LaTeX compilation errors and package conflicts. In all cases, the authors reviewed and verified for the correctness of the final content, experimental results, and code.

\bibliography{references}

@inproceedings{xiao2024streamingllm,
  title={Efficient streaming language models with attention sinks},
  author={Xiao, Guangxuan and Tian, Yuandong and Chen, Beidi and Han, Song and Lewis, Mike},
  booktitle={International Conference on Learning Representations},
  volume={2024},
  pages={21875--21895},
  year={2024}
}

@article{sun2025compressionvalleys,
  author       = {Enrique Queipo{-}de{-}Llano and
                  Alvaro Arroyo and
                  Federico Barbero and
                  Xiaowen Dong and
                  Michael M. Bronstein and
                  Yann LeCun and
                  Ravid Shwartz{-}Ziv},
  title        = {Attention Sinks and Compression Valleys in LLMs are Two Sides of the
                  Same Coin},
  journal      = {CoRR},
  volume       = {abs/2510.06477},
  year         = {2025},
  url          = {https://doi.org/10.48550/arXiv.2510.06477},
  doi          = {10.48550/ARXIV.2510.06477},
  eprinttype   = {arXiv},
  eprint       = {2510.06477},
  bibsource    = {dblp computer science bibliography, https://dblp.org}
}

@article{ganesh2025historyswapping,
  title={Whose Narrative is it Anyway? A KV Cache Manipulation Attack},
  author={Ganesh, Mukkesh and Iyer, Kaushik and Ananthan, Arun Baalaaji Sankar},
  journal={arXiv preprint arXiv:2511.12752},
  year={2025}
}

@article{cai2024pyramidkv,
  author       = {Zefan Cai and
                  Yichi Zhang and
                  Bofei Gao and
                  Yuliang Liu and
                  Tianyu Liu and
                  Keming Lu and
                  Wayne Xiong and
                  Yue Dong and
                  Baobao Chang and
                  Junjie Hu and
                  Wen Xiao},
  title        = {PyramidKV: Dynamic {KV} Cache Compression based on Pyramidal Information
                  Funneling},
  journal      = {CoRR},
  volume       = {abs/2406.02069},
  year         = {2024},
  url          = {https://doi.org/10.48550/arXiv.2406.02069},
  doi          = {10.48550/ARXIV.2406.02069},
  eprinttype   = {arXiv},
  eprint       = {2406.02069},
  bibsource    = {dblp computer science bibliography, https://dblp.org}
}

@inproceedings{li2024snapkv,
  author       = {Yuhong Li and
                  Yingbing Huang and
                  Bowen Yang and
                  Bharat Venkitesh and
                  Acyr Locatelli and
                  Hanchen Ye and
                  Tianle Cai and
                  Patrick Lewis and
                  Deming Chen},
  editor       = {Amir Globersons and
                  Lester Mackey and
                  Danielle Belgrave and
                  Angela Fan and
                  Ulrich Paquet and
                  Jakub M. Tomczak and
                  Cheng Zhang},
  title        = {SnapKV: {LLM} Knows What You are Looking for Before Generation},
  booktitle    = {Advances in Neural Information Processing Systems 37: Annual Conference
                  on Neural Information Processing Systems 2024, NeurIPS 2024, Vancouver,
                  BC, Canada, December 10 - 15, 2024},
  year         = {2024},
  url          = {http://papers.nips.cc/paper\_files/paper/2024/hash/28ab418242603e0f7323e54185d19bde-Abstract-Conference.html},
  bibsource    = {dblp computer science bibliography, https://dblp.org}
}

@article{zhang2023h2o,
  title={H2o: Heavy-hitter oracle for efficient generative inference of large language models},
  author  = {Zhang, Zhenyu and Sheng, Ying and Zhou, Tianyi and Chen, Tianlong and Zheng, Lianmin and Cai, Ruisi and Song, Zhao and Tian, Yuandong and R{\'e}, Christopher and Barrett, Clark and Wang, Zhangyang and Chen, Beidi},
  booktitle = {Advances in Neural Information Processing Systems},
  journal={Advances in Neural Information Processing Systems},
  volume={36},
  pages={34661--34710},
  year={2023}
}

@inproceedings{ge2024fastgen,
  author       = {Suyu Ge and
                  Yunan Zhang and
                  Liyuan Liu and
                  Minjia Zhang and
                  Jiawei Han and
                  Jianfeng Gao},
  title        = {Model Tells You What to Discard: Adaptive {KV} Cache Compression for
                  LLMs},
  booktitle    = {The Twelfth International Conference on Learning Representations,
                  {ICLR} 2024, Vienna, Austria, May 7-11, 2024},
  publisher    = {OpenReview.net},
  year         = {2024},
  url          = {https://openreview.net/forum?id=uNrFpDPMyo},
  bibsource    = {dblp computer science bibliography, https://dblp.org}
}

@inproceedings{liu2024scissorhands,
  author       = {Zichang Liu and
                  Aditya Desai and
                  Fangshuo Liao and
                  Weitao Wang and
                  Victor Xie and
                  Zhaozhuo Xu and
                  Anastasios Kyrillidis and
                  Anshumali Shrivastava},
  editor       = {Alice Oh and
                  Tristan Naumann and
                  Amir Globerson and
                  Kate Saenko and
                  Moritz Hardt and
                  Sergey Levine},
  title        = {Scissorhands: Exploiting the Persistence of Importance Hypothesis
                  for {LLM} {KV} Cache Compression at Test Time},
  booktitle    = {Advances in Neural Information Processing Systems 36: Annual Conference
                  on Neural Information Processing Systems 2023, NeurIPS 2023, New Orleans,
                  LA, USA, December 10 - 16, 2023},
  year         = {2023},
  url          = {http://papers.nips.cc/paper\_files/paper/2023/hash/a452a7c6c463e4ae8fbdc614c6e983e6-Abstract-Conference.html},
  bibsource    = {dblp computer science bibliography, https://dblp.org}
}

@article{olsson2022inductionheads,
  title={In-context learning and induction heads},
author={Catherine Olsson and Nelson Elhage and Neel Nanda and Nicholas Joseph and Nova DasSarma and Tom Henighan and Ben Mann and Amanda Askell and Yuntao Bai and Anna Chen and Tom Conerly and Dawn Drain and Deep Ganguli and Zac Hatfield-Dodds and Danny Hernandez and Scott Johnston and Andy Jones and Jackson Kernion and Liane Lovitt and Kamal Ndousse and Dario Amodei and Tom Brown and Jack Clark and Jared Kaplan and Sam McCandlish and Chris Olah},  journal={arXiv preprint arXiv:2209.11895},
  year={2022}
}

@inproceedings{hendel2023taskvectors,
  author       = {Roee Hendel and
                  Mor Geva and
                  Amir Globerson},
  editor       = {Houda Bouamor and
                  Juan Pino and
                  Kalika Bali},
  title        = {In-Context Learning Creates Task Vectors},
  booktitle    = {Findings of the Association for Computational Linguistics: {EMNLP}
                  2023, Singapore, December 6-10, 2023},
  series       = {Findings of {ACL}},
  volume       = {{EMNLP} 2023},
  pages        = {9318--9333},
  publisher    = {Association for Computational Linguistics},
  year         = {2023},
  url          = {https://doi.org/10.18653/v1/2023.findings-emnlp.624},
  doi          = {10.18653/V1/2023.FINDINGS-EMNLP.624},
  bibsource    = {dblp computer science bibliography, https://dblp.org}
}

@inproceedings{todd2024functionvectors,
  author       = {Eric Todd and
                  Millicent L. Li and
                  Arnab Sen Sharma and
                  Aaron Mueller and
                  Byron C. Wallace and
                  David Bau},
  title        = {Function Vectors in Large Language Models},
  booktitle    = {The Twelfth International Conference on Learning Representations,
                  {ICLR} 2024, Vienna, Austria, May 7-11, 2024},
  publisher    = {OpenReview.net},
  year         = {2024},
  url          = {https://openreview.net/forum?id=AwyxtyMwaG},
  bibsource    = {dblp computer science bibliography, https://dblp.org}
}

@article{cobbe2021gsm8k,
  title={Training verifiers to solve math word problems},
author={Karl Cobbe and Vineet Kosaraju and Mohammad Bavarian and Mark Chen and Heewoo Jun and Lukasz Kaiser and Matthias Plappert and Jerry Tworek and Jacob Hilton and Reiichiro Nakano and Christopher Hesse and John Schulman},
  journal={https://arxiv.org/abs/2110.14168},
  volume={9},
  year={2021}
}

@article{austin2021mbpp,
       title={Program Synthesis with Large Language Models}, 
      author={Jacob Austin and Augustus Odena and Maxwell Nye and Maarten Bosma and Henryk Michalewski and David Dohan and Ellen Jiang and Carrie Cai and Michael Terry and Quoc Le and Charles Sutton},
      year={2021},
      eprint={2108.07732},
      archivePrefix={arXiv},
      primaryClass={cs.PL},
      url={https://arxiv.org/abs/2108.07732}, 
}

@article{chen2021humaneval,
  title={Evaluating large language models trained on code},
      author={Mark Chen and Jerry Tworek and Heewoo Jun and Qiming Yuan and Henrique Ponde de Oliveira Pinto and Jared Kaplan and Harri Edwards and Yuri Burda and Nicholas Joseph and Greg Brockman and Alex Ray and Raul Puri and Gretchen Krueger and Michael Petrov and Heidy Khlaaf and Girish Sastry and Pamela Mishkin and Brooke Chan and Scott Gray and Nick Ryder and Mikhail Pavlov and Alethea Power and Lukasz Kaiser and Mohammad Bavarian and Clemens Winter and Philippe Tillet and Felipe Petroski Such and Dave Cummings and Matthias Plappert and Fotios Chantzis and Elizabeth Barnes and Ariel Herbert-Voss and William Hebgen Guss and Alex Nichol and Alex Paino and Nikolas Tezak and Jie Tang and Igor Babuschkin and Suchir Balaji and Shantanu Jain and William Saunders and Christopher Hesse and Andrew N. Carr and Jan Leike and Josh Achiam and Vedant Misra and Evan Morikawa and Alec Radford and Matthew Knight and Miles Brundage and Mira Murati and Katie Mayer and Peter Welinder and Bob McGrew and Dario Amodei and Sam McCandlish and Ilya Sutskever and Wojciech Zaremba},  
      journal={arXiv preprint arXiv:2107.03374},
  year={2021}
}

@article{qwen3,
  title={Qwen3 technical report},
  author={An Yang and Anfeng Li and Baosong Yang and Beichen Zhang and Binyuan Hui and Bo Zheng and Bowen Yu and Chang Gao and Chengen Huang and Chenxu Lv and Chujie Zheng and Dayiheng Liu and Fan Zhou and Fei Huang and Feng Hu and Hao Ge and Haoran Wei and Huan Lin and Jialong Tang and Jian Yang and Jianhong Tu and Jianwei Zhang and Jianxin Yang and Jiaxi Yang and Jing Zhou and Jingren Zhou and Junyang Lin and Kai Dang and Keqin Bao and Kexin Yang and Le Yu and Lianghao Deng and Mei Li and Mingfeng Xue and Mingze Li and Pei Zhang and Peng Wang and Qin Zhu and Rui Men and Ruize Gao and Shixuan Liu and Shuang Luo and Tianhao Li and Tianyi Tang and Wenbiao Yin and Xingzhang Ren and Xinyu Wang and Xinyu Zhang and Xuancheng Ren and Yang Fan and Yang Su and Yichang Zhang and Yinger Zhang and Yu Wan and Yuqiong Liu and Zekun Wang and Zeyu Cui and Zhenru Zhang and Zhipeng Zhou and Zihan Qiu},
  journal={arXiv preprint arXiv:2505.09388},
  year={2025}
}

@article{gemma3,
  title={Gemma 3 technical report},
author={Gemma Team and Aishwarya Kamath and Johan Ferret and Shreya Pathak and Nino Vieillard and Ramona Merhej and Sarah Perrin and Tatiana Matejovicova and Alexandre Ramé and Morgane Rivière and Louis Rouillard and Thomas Mesnard and Geoffrey Cideron and Jean-bastien Grill and Sabela Ramos and Edouard Yvinec and Michelle Casbon and Etienne Pot and Ivo Penchev and Gaël Liu and Francesco Visin and Kathleen Kenealy and Lucas Beyer and Xiaohai Zhai and Anton Tsitsulin and Robert Busa-Fekete and Alex Feng and Noveen Sachdeva and Benjamin Coleman and Yi Gao and Basil Mustafa and Iain Barr and Emilio Parisotto and David Tian and Matan Eyal and Colin Cherry and Jan-Thorsten Peter and Danila Sinopalnikov and Surya Bhupatiraju and Rishabh Agarwal and Mehran Kazemi and Dan Malkin and Ravin Kumar and David Vilar and Idan Brusilovsky and Jiaming Luo and Andreas Steiner and Abe Friesen and Abhanshu Sharma and Abheesht Sharma and Adi Mayrav Gilady and Adrian Goedeckemeyer and Alaa Saade and Alex Feng and Alexander Kolesnikov and Alexei Bendebury and Alvin Abdagic and Amit Vadi and András György and André Susano Pinto and Anil Das and Ankur Bapna and Antoine Miech and Antoine Yang and Antonia Paterson and Ashish Shenoy and Ayan Chakrabarti and Bilal Piot and Bo Wu and Bobak Shahriari and Bryce Petrini and Charlie Chen and Charline Le Lan and Christopher A. Choquette-Choo and CJ Carey and Cormac Brick and Daniel Deutsch and Danielle Eisenbud and Dee Cattle and Derek Cheng and Dimitris Paparas and Divyashree Shivakumar Sreepathihalli and Doug Reid and Dustin Tran and Dustin Zelle and Eric Noland and Erwin Huizenga and Eugene Kharitonov and Frederick Liu and Gagik Amirkhanyan and Glenn Cameron and Hadi Hashemi and Hanna Klimczak-Plucińska and Harman Singh and Harsh Mehta and Harshal Tushar Lehri and Hussein Hazimeh and Ian Ballantyne and Idan Szpektor and Ivan Nardini and Jean Pouget-Abadie and Jetha Chan and Joe Stanton and John Wieting and Jonathan Lai and Jordi Orbay and Joseph Fernandez and Josh Newlan and Ju-yeong Ji and Jyotinder Singh and Kat Black and Kathy Yu and Kevin Hui and Kiran Vodrahalli and Klaus Greff and Linhai Qiu and Marcella Valentine and Marina Coelho and Marvin Ritter and Matt Hoffman and Matthew Watson and Mayank Chaturvedi and Michael Moynihan and Min Ma and Nabila Babar and Natasha Noy and Nathan Byrd and Nick Roy and Nikola Momchev and Nilay Chauhan and Noveen Sachdeva and Oskar Bunyan and Pankil Botarda and Paul Caron and Paul Kishan Rubenstein and Phil Culliton and Philipp Schmid and Pier Giuseppe Sessa and Pingmei Xu and Piotr Stanczyk and Pouya Tafti and Rakesh Shivanna and Renjie Wu and Renke Pan and Reza Rokni and Rob Willoughby and Rohith Vallu and Ryan Mullins and Sammy Jerome and Sara Smoot and Sertan Girgin and Shariq Iqbal and Shashir Reddy and Shruti Sheth and Siim Põder and Sijal Bhatnagar and Sindhu Raghuram Panyam and Sivan Eiger and Susan Zhang and Tianqi Liu and Trevor Yacovone and Tyler Liechty and Uday Kalra and Utku Evci and Vedant Misra and Vincent Roseberry and Vlad Feinberg and Vlad Kolesnikov and Woohyun Han and Woosuk Kwon and Xi Chen and Yinlam Chow and Yuvein Zhu and Zichuan Wei and Zoltan Egyed and Victor Cotruta and Minh Giang and Phoebe Kirk and Anand Rao and Kat Black and Nabila Babar and Jessica Lo and Erica Moreira and Luiz Gustavo Martins and Omar Sanseviero and Lucas Gonzalez and Zach Gleicher and Tris Warkentin and Vahab Mirrokni and Evan Senter and Eli Collins and Joelle Barral and Zoubin Ghahramani and Raia Hadsell and Yossi Matias and D. Sculley and Slav Petrov and Noah Fiedel and Noam Shazeer and Oriol Vinyals and Jeff Dean and Demis Hassabis and Koray Kavukcuoglu and Clement Farabet and Elena Buchatskaya and Jean-Baptiste Alayrac and Rohan Anil and Dmitry and Lepikhin and Sebastian Borgeaud and Olivier Bachem and Armand Joulin and Alek Andreev and Cassidy Hardin and Robert Dadashi and Léonard Hussenot},  
    journal={arXiv preprint arXiv:2503.19786},
  volume={4},
  year={2025},
  publisher={ArXiv}
}

@article{grattafiori2024llama3,
  title={The llama 3 herd of models},
    author={Aaron Grattafiori and Abhimanyu Dubey and Abhinav Jauhri and Abhinav Pandey and Abhishek Kadian and Ahmad Al-Dahle and Aiesha Letman and Akhil Mathur and Alan Schelten and Alex Vaughan and Amy Yang and Angela Fan and Anirudh Goyal and Anthony Hartshorn and Aobo Yang and Archi Mitra and Archie Sravankumar and Artem Korenev and Arthur Hinsvark and Arun Rao and Aston Zhang and Aurelien Rodriguez and Austen Gregerson and Ava Spataru and Baptiste Roziere and Bethany Biron and Binh Tang and Bobbie Chern and Charlotte Caucheteux and Chaya Nayak and Chloe Bi and Chris Marra and Chris McConnell and Christian Keller and Christophe Touret and Chunyang Wu and Corinne Wong and Cristian Canton Ferrer and Cyrus Nikolaidis and Damien Allonsius and Daniel Song and Danielle Pintz and Danny Livshits and Danny Wyatt and David Esiobu and Dhruv Choudhary and Dhruv Mahajan and Diego Garcia-Olano and Diego Perino and Dieuwke Hupkes and Egor Lakomkin and Ehab AlBadawy and Elina Lobanova and Emily Dinan and Eric Michael Smith and Filip Radenovic and Francisco Guzmán and Frank Zhang and Gabriel Synnaeve and Gabrielle Lee and Georgia Lewis Anderson and Govind Thattai and Graeme Nail and Gregoire Mialon and Guan Pang and Guillem Cucurell and Hailey Nguyen and Hannah Korevaar and Hu Xu and Hugo Touvron and Iliyan Zarov and Imanol Arrieta Ibarra and Isabel Kloumann and Ishan Misra and Ivan Evtimov and Jack Zhang and Jade Copet and Jaewon Lee and Jan Geffert and Jana Vranes and Jason Park and Jay Mahadeokar and Jeet Shah and Jelmer van der Linde and Jennifer Billock and Jenny Hong and Jenya Lee and Jeremy Fu and Jianfeng Chi and Jianyu Huang and Jiawen Liu and Jie Wang and Jiecao Yu and Joanna Bitton and Joe Spisak and Jongsoo Park and Joseph Rocca and Joshua Johnstun and Joshua Saxe and Junteng Jia and Kalyan Vasuden Alwala and Karthik Prasad and Kartikeya Upasani and Kate Plawiak and Ke Li and Kenneth Heafield and Kevin Stone and Khalid El-Arini and Krithika Iyer and Kshitiz Malik and Kuenley Chiu and Kunal Bhalla and Kushal Lakhotia and Lauren Rantala-Yeary and Laurens van der Maaten and Lawrence Chen and Liang Tan and Liz Jenkins and Louis Martin and Lovish Madaan and Lubo Malo and Lukas Blecher and Lukas Landzaat and Luke de Oliveira and Madeline Muzzi and Mahesh Pasupuleti and Mannat Singh and Manohar Paluri and Marcin Kardas and Maria Tsimpoukelli and Mathew Oldham and Mathieu Rita and Maya Pavlova and Melanie Kambadur and Mike Lewis and Min Si and Mitesh Kumar Singh and Mona Hassan and Naman Goyal and Narjes Torabi and Nikolay Bashlykov and Nikolay Bogoychev and Niladri Chatterji and Ning Zhang and Olivier Duchenne and Onur Çelebi and Patrick Alrassy and Pengchuan Zhang and Pengwei Li and Petar Vasic and Peter Weng and Prajjwal Bhargava and Pratik Dubal and Praveen Krishnan and Punit Singh Koura and Puxin Xu and Qing He and Qingxiao Dong and Ragavan Srinivasan and Raj Ganapathy and Ramon Calderer and Ricardo Silveira Cabral and Robert Stojnic and Roberta Raileanu and Rohan Maheswari and Rohit Girdhar and Rohit Patel and Romain Sauvestre and Ronnie Polidoro and Roshan Sumbaly and Ross Taylor and Ruan Silva and Rui Hou and Rui Wang and Saghar Hosseini and Sahana Chennabasappa and Sanjay Singh and Sean Bell and Seohyun Sonia Kim and Sergey Edunov and Shaoliang Nie and Sharan Narang and Sharath Raparthy and Sheng Shen and Shengye Wan and Shruti Bhosale and Shun Zhang and Simon Vandenhende and Soumya Batra and Spencer Whitman and Sten Sootla and Stephane Collot and Suchin Gururangan and Sydney Borodinsky and Tamar Herman and Tara Fowler and Tarek Sheasha and Thomas Georgiou and Thomas Scialom and Tobias Speckbacher and Todor Mihaylov and Tong Xiao and Ujjwal Karn and Vedanuj Goswami and Vibhor Gupta and Vignesh Ramanathan and Viktor Kerkez and Vincent Gonguet and Virginie Do and Vish Vogeti and Vítor Albiero and Vladan Petrovic and Weiwei Chu and Wenhan Xiong and Wenyin Fu and Whitney Meers and Xavier Martinet and Xiaodong Wang and Xiaofang Wang and Xiaoqing Ellen Tan and Xide Xia and Xinfeng Xie and Xuchao Jia and Xuewei Wang and Yaelle Goldschlag and Yashesh Gaur and Yasmine Babaei and Yi Wen and Yiwen Song and Yuchen Zhang and Yue Li and Yuning Mao and Zacharie Delpierre Coudert and Zheng Yan and Zhengxing Chen and Zoe Papakipos and Aaditya Singh and Aayushi Srivastava and Abha Jain and Adam Kelsey and Adam Shajnfeld and Adithya Gangidi and Adolfo Victoria and Ahuva Goldstand and Ajay Menon and Ajay Sharma and Alex Boesenberg and Alexei Baevski and Allie Feinstein and Amanda Kallet and Amit Sangani and Amos Teo and Anam Yunus and Andrei Lupu and Andres Alvarado and Andrew Caples and Andrew Gu and Andrew Ho and Andrew Poulton and Andrew Ryan and Ankit Ramchandani and Annie Dong and Annie Franco and Anuj Goyal and Aparajita Saraf and Arkabandhu Chowdhury and Ashley Gabriel and Ashwin Bharambe and Assaf Eisenman and Azadeh Yazdan and Beau James and Ben Maurer and Benjamin Leonhardi and Bernie Huang and Beth Loyd and Beto De Paola and Bhargavi Paranjape and Bing Liu and Bo Wu and Boyu Ni and Braden Hancock and Bram Wasti and Brandon Spence and Brani Stojkovic and Brian Gamido and Britt Montalvo and Carl Parker and Carly Burton and Catalina Mejia and Ce Liu and Changhan Wang and Changkyu Kim and Chao Zhou and Chester Hu and Ching-Hsiang Chu and Chris Cai and Chris Tindal and Christoph Feichtenhofer and Cynthia Gao and Damon Civin and Dana Beaty and Daniel Kreymer and Daniel Li and David Adkins and David Xu and Davide Testuggine and Delia David and Devi Parikh and Diana Liskovich and Didem Foss and Dingkang Wang and Duc Le and Dustin Holland and Edward Dowling and Eissa Jamil and Elaine Montgomery and Eleonora Presani and Emily Hahn and Emily Wood and Eric-Tuan Le and Erik Brinkman and Esteban Arcaute and Evan Dunbar and Evan Smothers and Fei Sun and Felix Kreuk and Feng Tian and Filippos Kokkinos and Firat Ozgenel and Francesco Caggioni and Frank Kanayet and Frank Seide and Gabriela Medina Florez and Gabriella Schwarz and Gada Badeer and Georgia Swee and Gil Halpern and Grant Herman and Grigory Sizov and Guangyi and Zhang and Guna Lakshminarayanan and Hakan Inan and Hamid Shojanazeri and Han Zou and Hannah Wang and Hanwen Zha and Haroun Habeeb and Harrison Rudolph and Helen Suk and Henry Aspegren and Hunter Goldman and Hongyuan Zhan and Ibrahim Damlaj and Igor Molybog and Igor Tufanov and Ilias Leontiadis and Irina-Elena Veliche and Itai Gat and Jake Weissman and James Geboski and James Kohli and Janice Lam and Japhet Asher and Jean-Baptiste Gaya and Jeff Marcus and Jeff Tang and Jennifer Chan and Jenny Zhen and Jeremy Reizenstein and Jeremy Teboul and Jessica Zhong and Jian Jin and Jingyi Yang and Joe Cummings and Jon Carvill and Jon Shepard and Jonathan McPhie and Jonathan Torres and Josh Ginsburg and Junjie Wang and Kai Wu and Kam Hou U and Karan Saxena and Kartikay Khandelwal and Katayoun Zand and Kathy Matosich and Kaushik Veeraraghavan and Kelly Michelena and Keqian Li and Kiran Jagadeesh and Kun Huang and Kunal Chawla and Kyle Huang and Lailin Chen and Lakshya Garg and Lavender A and Leandro Silva and Lee Bell and Lei Zhang and Liangpeng Guo and Licheng Yu and Liron Moshkovich and Luca Wehrstedt and Madian Khabsa and Manav Avalani and Manish Bhatt and Martynas Mankus and Matan Hasson and Matthew Lennie and Matthias Reso and Maxim Groshev and Maxim Naumov and Maya Lathi and Meghan Keneally and Miao Liu and Michael L. Seltzer and Michal Valko and Michelle Restrepo and Mihir Patel and Mik Vyatskov and Mikayel Samvelyan and Mike Clark and Mike Macey and Mike Wang and Miquel Jubert Hermoso and Mo Metanat and Mohammad Rastegari and Munish Bansal and Nandhini Santhanam and Natascha Parks and Natasha White and Navyata Bawa and Nayan Singhal and Nick Egebo and Nicolas Usunier and Nikhil Mehta and Nikolay Pavlovich Laptev and Ning Dong and Norman Cheng and Oleg Chernoguz and Olivia Hart and Omkar Salpekar and Ozlem Kalinli and Parkin Kent and Parth Parekh and Paul Saab and Pavan Balaji and Pedro Rittner and Philip Bontrager and Pierre Roux and Piotr Dollar and Polina Zvyagina and Prashant Ratanchandani and Pritish Yuvraj and Qian Liang and Rachad Alao and Rachel Rodriguez and Rafi Ayub and Raghotham Murthy and Raghu Nayani and Rahul Mitra and Rangaprabhu Parthasarathy and Raymond Li and Rebekkah Hogan and Robin Battey and Rocky Wang and Russ Howes and Ruty Rinott and Sachin Mehta and Sachin Siby and Sai Jayesh Bondu and Samyak Datta and Sara Chugh and Sara Hunt and Sargun Dhillon and Sasha Sidorov and Satadru Pan and Saurabh Mahajan and Saurabh Verma and Seiji Yamamoto and Sharadh Ramaswamy and Shaun Lindsay and Shaun Lindsay and Sheng Feng and Shenghao Lin and Shengxin Cindy Zha and Shishir Patil and Shiva Shankar and Shuqiang Zhang and Shuqiang Zhang and Sinong Wang and Sneha Agarwal and Soji Sajuyigbe and Soumith Chintala and Stephanie Max and Stephen Chen and Steve Kehoe and Steve Satterfield and Sudarshan Govindaprasad and Sumit Gupta and Summer Deng and Sungmin Cho and Sunny Virk and Suraj Subramanian and Sy Choudhury and Sydney Goldman and Tal Remez and Tamar Glaser and Tamara Best and Thilo Koehler and Thomas Robinson and Tianhe Li and Tianjun Zhang and Tim Matthews and Timothy Chou and Tzook Shaked and Varun Vontimitta and Victoria Ajayi and Victoria Montanez and Vijai Mohan and Vinay Satish Kumar and Vishal Mangla and Vlad Ionescu and Vlad Poenaru and Vlad Tiberiu Mihailescu and Vladimir Ivanov and Wei Li and Wenchen Wang and Wenwen Jiang and Wes Bouaziz and Will Constable and Xiaocheng Tang and Xiaojian Wu and Xiaolan Wang and Xilun Wu and Xinbo Gao and Yaniv Kleinman and Yanjun Chen and Ye Hu and Ye Jia and Ye Qi and Yenda Li and Yilin Zhang and Ying Zhang and Yossi Adi and Youngjin Nam and Yu and Wang and Yu Zhao and Yuchen Hao and Yundi Qian and Yunlu Li and Yuzi He and Zach Rait and Zachary DeVito and Zef Rosnbrick and Zhaoduo Wen and Zhenyu Yang and Zhiwei Zhao and Zhiyu Ma},  
    journal={arXiv preprint arXiv:2407.21783},
  year={2024}
}

@article{liu2024lostmiddle,
  title={Lost in the middle: How language models use long contexts},
  author={Liu, Nelson F and Lin, Kevin and Hewitt, John and Paranjape, Ashwin and Bevilacqua, Michele and Petroni, Fabio and Liang, Percy},
  journal={arXiv preprint arXiv:2307.03172},
  year={2023}
}

@article{zhou2023ifeval,
  author       = {Jeffrey Zhou and
                  Tianjian Lu and
                  Swaroop Mishra and
                  Siddhartha Brahma and
                  Sujoy Basu and
                  Yi Luan and
                  Denny Zhou and
                  Le Hou},
  title        = {Instruction-Following Evaluation for Large Language Models},
  journal      = {CoRR},
  volume       = {abs/2311.07911},
  year         = {2023},
  url          = {https://doi.org/10.48550/arXiv.2311.07911},
  doi          = {10.48550/ARXIV.2311.07911},
  eprinttype   = {arXiv},
  eprint       = {2311.07911},
  bibsource    = {dblp computer science bibliography, https://dblp.org}
}

@inproceedings{lifbench,
    title = "{LIFB}ench: Evaluating the Instruction Following Performance and Stability of Large Language Models in Long-Context Scenarios",
    author = "Wu, Xiaodong  and
      Wang, Minhao  and
      Liu, Yichen  and
      Shi, Xiaoming  and
      Yan, He  and
      Xiangju, Lu  and
      Zhu, Junmin  and
      Zhang, Wei",
    editor = "Che, Wanxiang  and
      Nabende, Joyce  and
      Shutova, Ekaterina  and
      Pilehvar, Mohammad Taher",
    booktitle = "Proceedings of the 63rd Annual Meeting of the {A}ssociation for {C}omputational {L}inguistics (Volume 1: Long Papers)",
    month = jul,
    year = "2025",
    address = "Vienna, Austria",
    publisher = "Association for Computational Linguistics",
    url = "https://aclanthology.org/2025.acl-long.803/",
    doi = "10.18653/v1/2025.acl-long.803",
    pages = "16445--16468",
    ISBN = "979-8-89176-251-0"
}

\newpage
\appendix
\textbf{\Large Appendix}

\section{Datasets and Scoring}
\label{app:datasets}

We evaluate on four datasets, sampling 100 problems from each to keep the
sweep over layer cutoff and decoding step manageable. All generation is
greedy.

\textbf{GSM8K}~\citep{cobbe2021gsm8k} is a chain-of-thought arithmetic
benchmark; we score a completion by exact match on the final numeric answer.

\textbf{MBPP}~\citep{austin2021mbpp} and
\textbf{HumanEval}~\citep{chen2021humaneval} are Python code-completion
benchmarks; we score a completion by whether it passes the test suite shipped
with each problem.

\textbf{IFEval}~\citep{zhou2023ifeval} is an instruction-following benchmark
of prompts augmented with programmatically verifiable instructions; we score
a completion as correct only if every instruction it specifies is satisfied.

\textbf{LIFBench}~\citep{lifbench} is a long-context
  instruction-following benchmark of eleven tasks, each instantiated at multiple context lengths. We use the three length levels that fit the models' context window (${\sim}$4k/8k/16k tokens), sampling 150 examples per length stratified evenly over tasks. Completions are scored with the benchmark's
  official deterministic rubric judges, which assign partial credit per
  instruction component and normalize each example to $[0,1]$.



\section{Algorithmic-Donor Benchmark: Construction and Validation}
\label{app:algo-bench}

The algorithmic-donor benchmark contains nine algorithmic families:
search, sorting, graph traversal, shortest path, Fibonacci, maximum
subarray, prime generation, two-sum, and string matching. Each family
contains ten problems, and every problem is presented in two variants
that share the same input-output contract but require different algorithmic
strategies. This produces 90 problems and 180 prompt variants, each indexed
by a problem, algorithm class, and family triple $(p,c,f)$.

\paragraph{Algorithm taxonomy.}
We define a pair of algorithms for each family: linear and binary search,
bubble and merge sort, BFS and DFS, Dijkstra and Bellman-Ford, naive
recursion and bottom-up dynamic programming for Fibonacci, brute force and
Kadane's algorithm for maximum subarray, trial division and the Sieve of
Eratosthenes for prime generation, nested loops and a hash map for two-sum,
and naive matching and KMP for string matching. We refer to the two
algorithms in each pair as sides A and B.

\paragraph{Variant generation.}
We use Sonnet 4.5 with \texttt{max\_tokens}=4096 and
\texttt{temperature}=0.8 to generate ten concrete problems for each
algorithmic family. The system prompt is shown below.

\begin{promptbox}{Variant Generation System Prompt}
\small\ttfamily
You are a programming problem generator. Given an algorithm family
with algorithms A and B, generate exactly 10 diverse problem variants.
Each variant should be a different concrete problem (different domain,
data types, or scenario) that both algorithms can solve.\newline
\newline
For each variant, provide:\newline
1. A Python prompt asking for algorithm B\newline
2. A Python solution using algorithm A\newline
\newline
Return a JSON array of 10 objects, each with keys:\newline
- ``variant\_name'': short description\newline
- ``python\_b\_prompt'': one-sentence Python prompt explicitly naming
algorithm B\newline
- ``python\_a\_code'': complete Python function implementing algorithm
A\newline
\newline
Return ONLY valid JSON, no prose.
\label{prompt:system}
\end{promptbox}

The output is
constrained to a JSON array of ten objects per family. We retry each call
up to three times in the event of parsing or transport errors.

\paragraph{Constructing the paired variants.}
We use a second generation pass with the same model and
\texttt{temperature}=0.2 to construct the missing half of each pair.
Given the A-side implementation and the name of algorithm B, the model
produces a B-side implementation with the same function signature and an
A-side prompt for the same problem. Each problem therefore contains four
artifacts: an A-side prompt and implementation and a B-side prompt and
implementation.

\paragraph{Ground truth and scoring.}
Each problem includes a deterministic hidden test suite with five cases,
consisting of one edge case and four standard cases. The reference
implementations are checked for internal consistency, although the splice
experiments use only the prompts and test cases. We primarily score a
completion according to whether it passes the hidden test suite. We
separately use an LLM judge to determine whether the generated solution
follows the algorithm class requested by the target prompt.

\paragraph{Donor selection.}
For a target variant $(p,c,f)$, a \textsc{same-algo} donor uses a different
problem from the same family and algorithm class. A \textsc{diff-algo}
donor uses the alternative algorithm class from the same family, while a
\textsc{diff-family} donor comes from another algorithmic family. These
conditions progressively increase the distance between the donor content
and the target task.

\paragraph{Human validation.}
We manually audit a random sample of 40 prompt variants to validate the
generated problems, test suites, reference implementations, and
algorithm-class judge. All 40 problems, tests, and reference solutions are
correct. On the same sample, the LLM judge agrees with a human annotator on
the algorithm class for 39 of the 40 variants.

\section{Per-Model Phase Diagrams}
\label{sec:appendix-heatmaps}

\begin{figure*}[p]
\centering
\begin{subfigure}[t]{0.49\linewidth}\centering
\includegraphics[width=\linewidth]{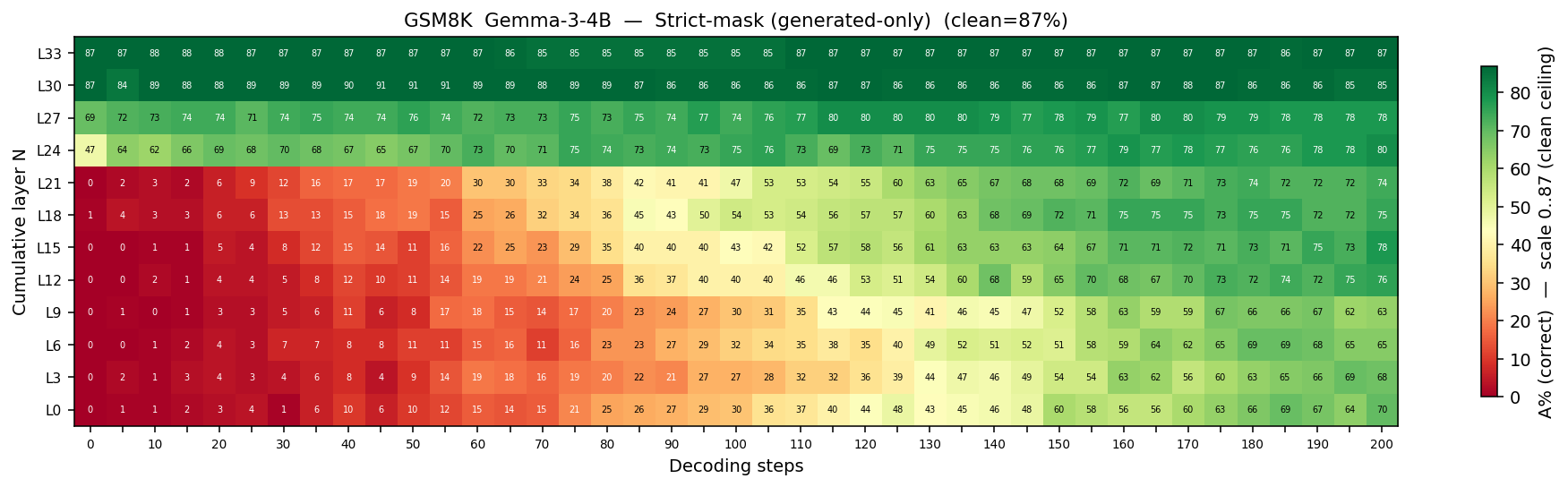}
\caption{Gemma-3-4B-IT, GSM8K, \textsc{zero}}
\end{subfigure}\hfill
\begin{subfigure}[t]{0.49\linewidth}\centering
\includegraphics[width=\linewidth]{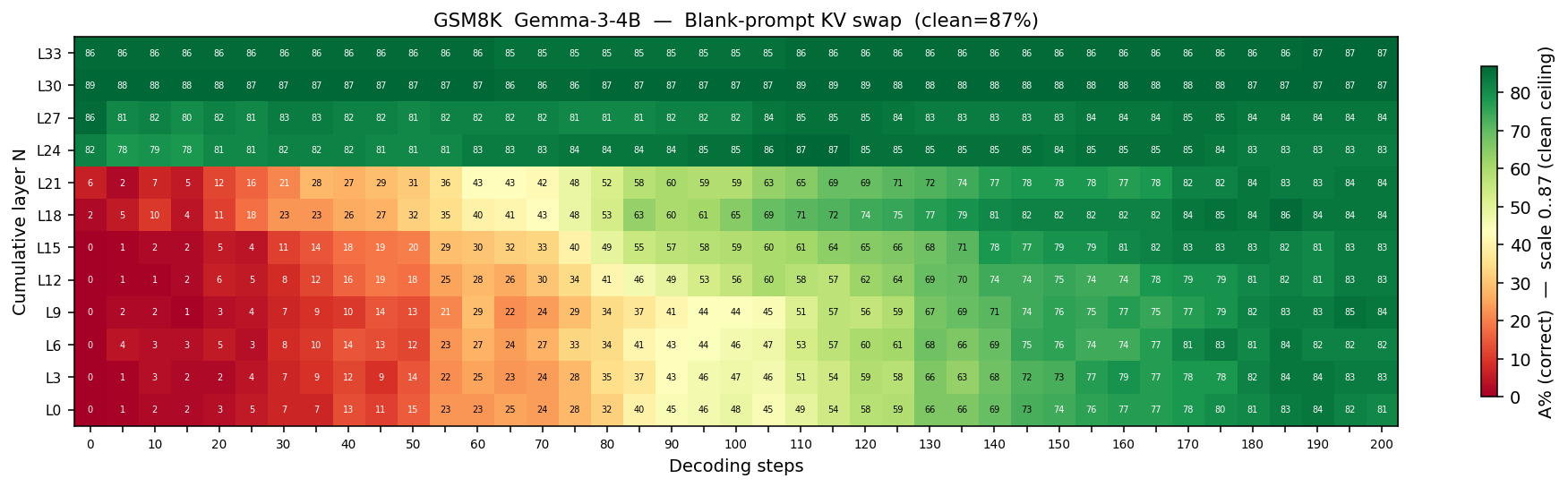}
\caption{Gemma-3-4B-IT, GSM8K, \textsc{blank}}
\end{subfigure}\\[0.4em]
\begin{subfigure}[t]{0.49\linewidth}\centering
\includegraphics[width=\linewidth]{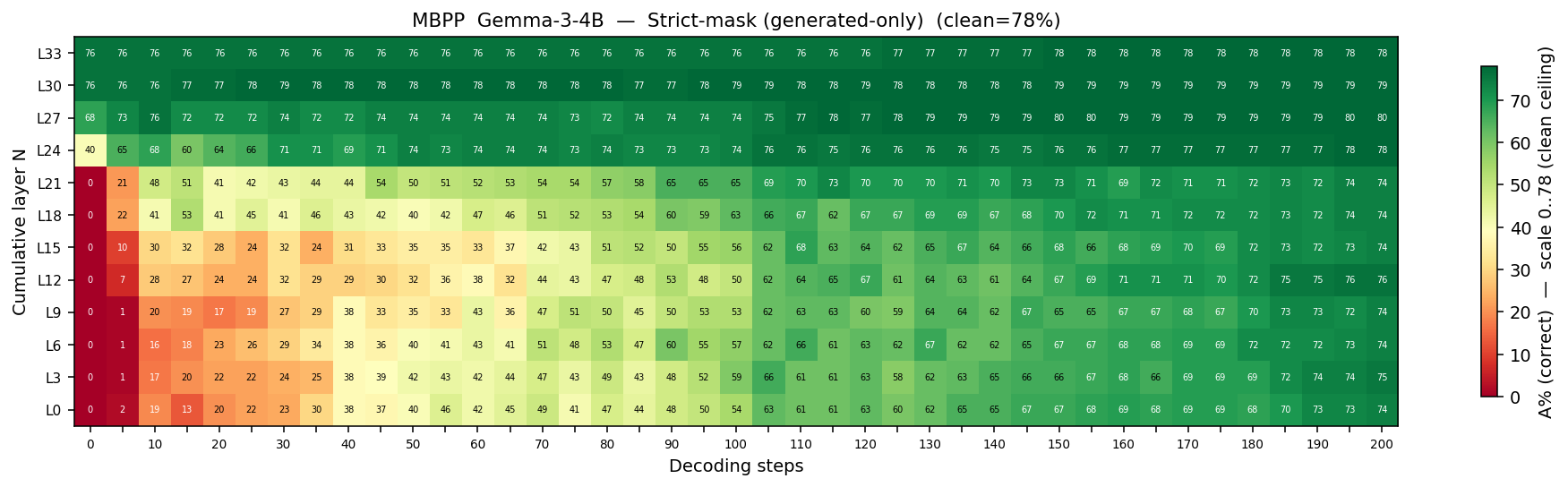}
\caption{Gemma-3-4B-IT, MBPP, \textsc{zero}}
\end{subfigure}\hfill
\begin{subfigure}[t]{0.49\linewidth}\centering
\includegraphics[width=\linewidth]{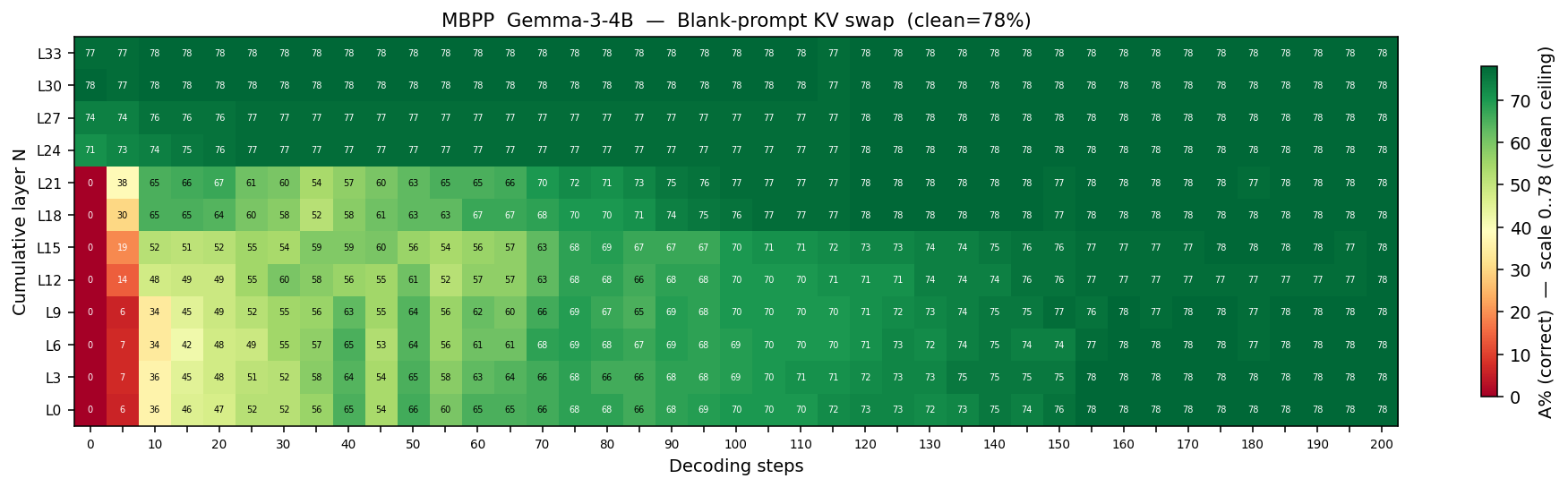}
\caption{Gemma-3-4B-IT, MBPP, \textsc{blank}}
\end{subfigure}\\[0.4em]
\begin{subfigure}[t]{0.49\linewidth}\centering
\includegraphics[width=\linewidth]{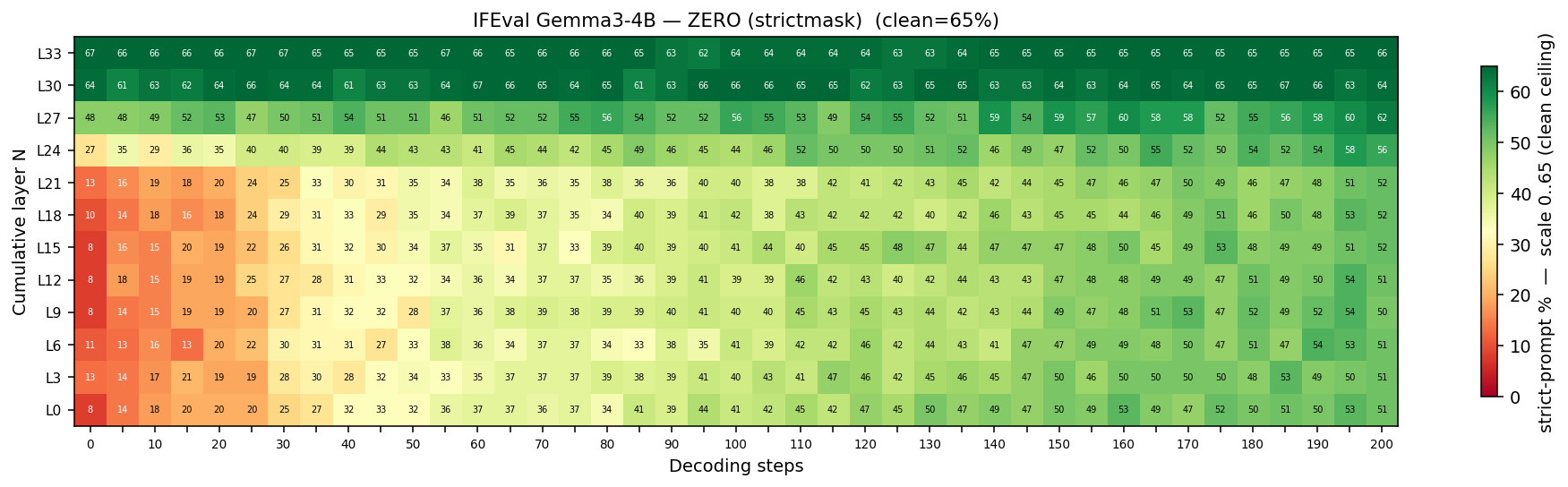}
\caption{Gemma-3-4B-IT, IFEval, \textsc{zero}}
\end{subfigure}\hfill
\begin{subfigure}[t]{0.49\linewidth}\centering
\includegraphics[width=\linewidth]{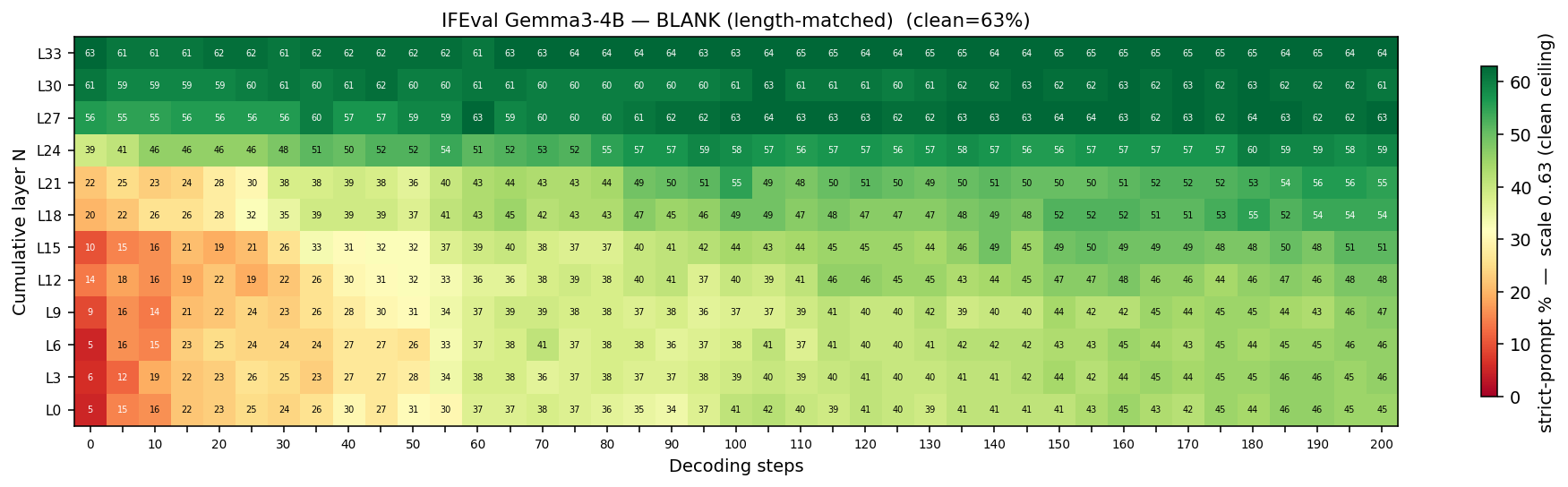}
\caption{Gemma-3-4B-IT, IFEval, \textsc{blank}}
\end{subfigure}\\[0.4em]
\begin{subfigure}[t]{0.49\linewidth}\centering
\includegraphics[width=\linewidth]{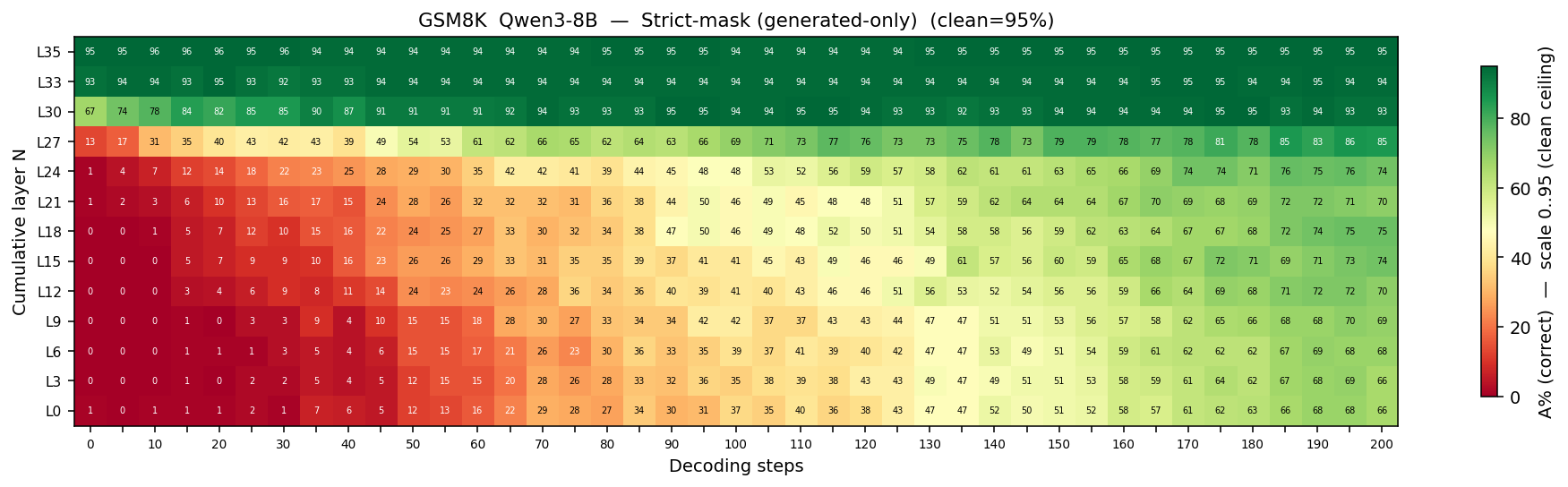}
\caption{Qwen3-8B, GSM8K, \textsc{zero}}
\end{subfigure}\hfill
\begin{subfigure}[t]{0.49\linewidth}\centering
\includegraphics[width=\linewidth]{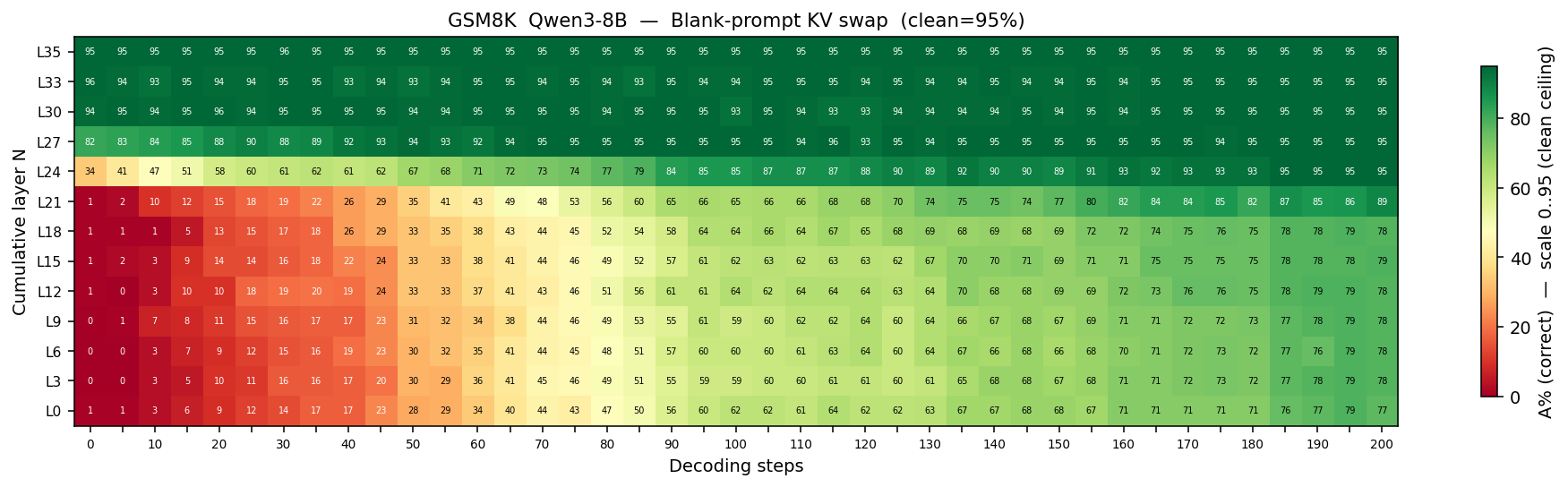}
\caption{Qwen3-8B, GSM8K, \textsc{blank}}
\end{subfigure}\\[0.4em]
\begin{subfigure}[t]{0.49\linewidth}\centering
\includegraphics[width=\linewidth]{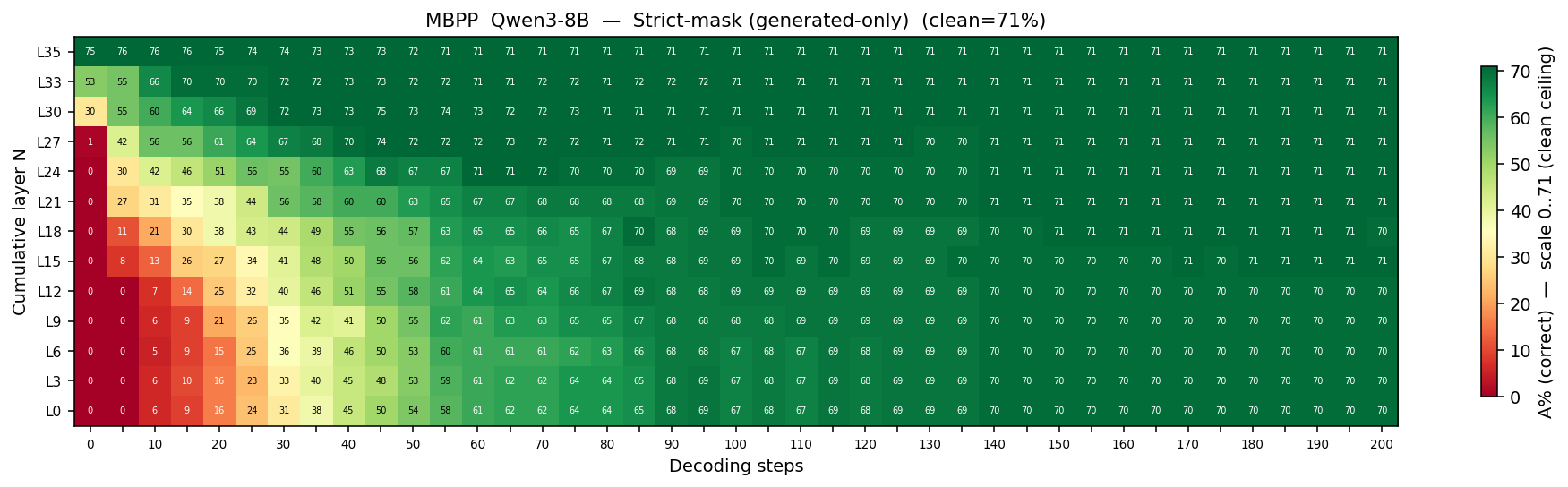}
\caption{Qwen3-8B, MBPP, \textsc{zero}}
\end{subfigure}\hfill
\begin{subfigure}[t]{0.49\linewidth}\centering
\includegraphics[width=\linewidth]{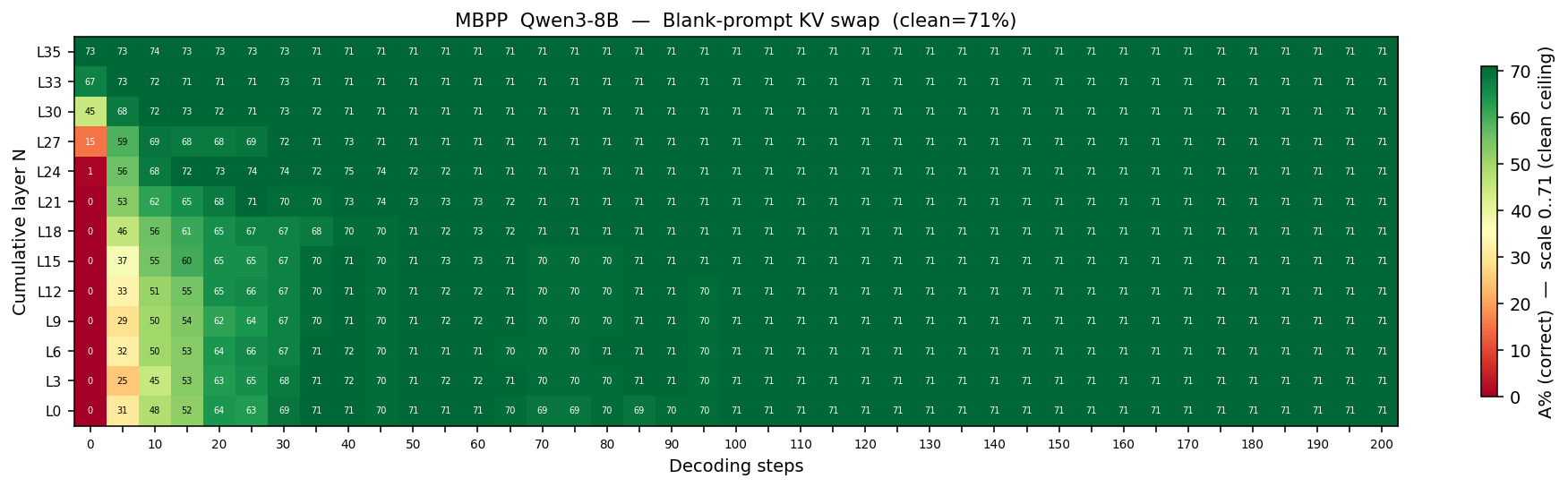}
\caption{Qwen3-8B, MBPP, \textsc{blank}}
\end{subfigure}\\[0.4em]
\begin{subfigure}[t]{0.49\linewidth}\centering
\includegraphics[width=\linewidth]{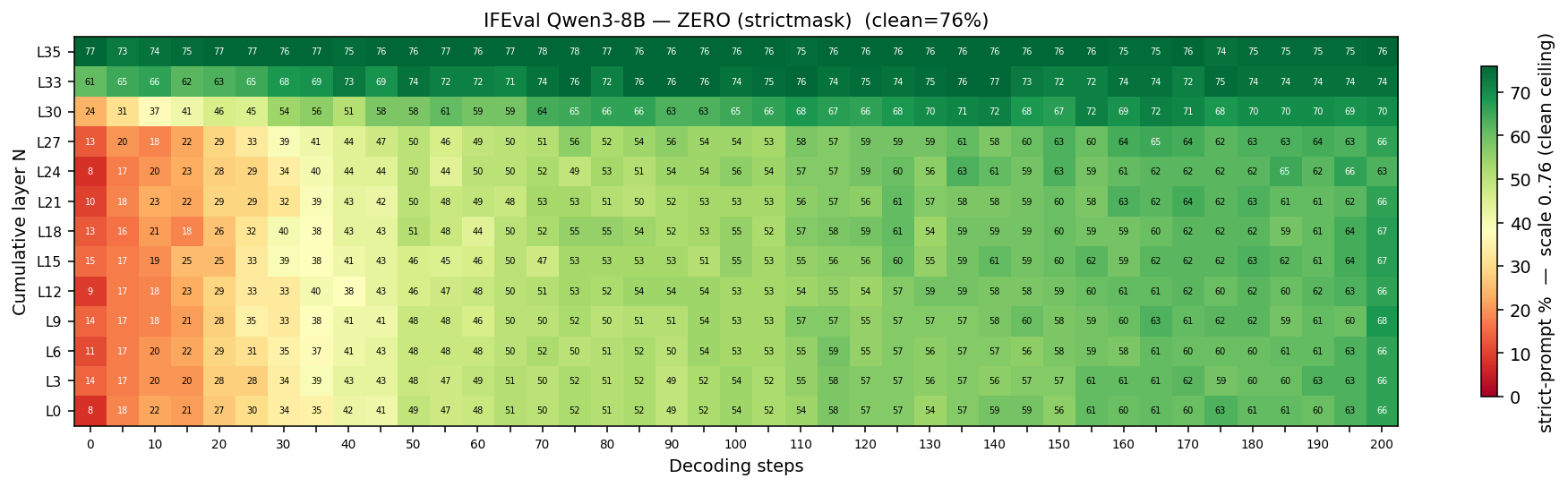}
\caption{Qwen3-8B, IFEval, \textsc{zero}}
\end{subfigure}\hfill
\begin{subfigure}[t]{0.49\linewidth}\centering
\includegraphics[width=\linewidth]{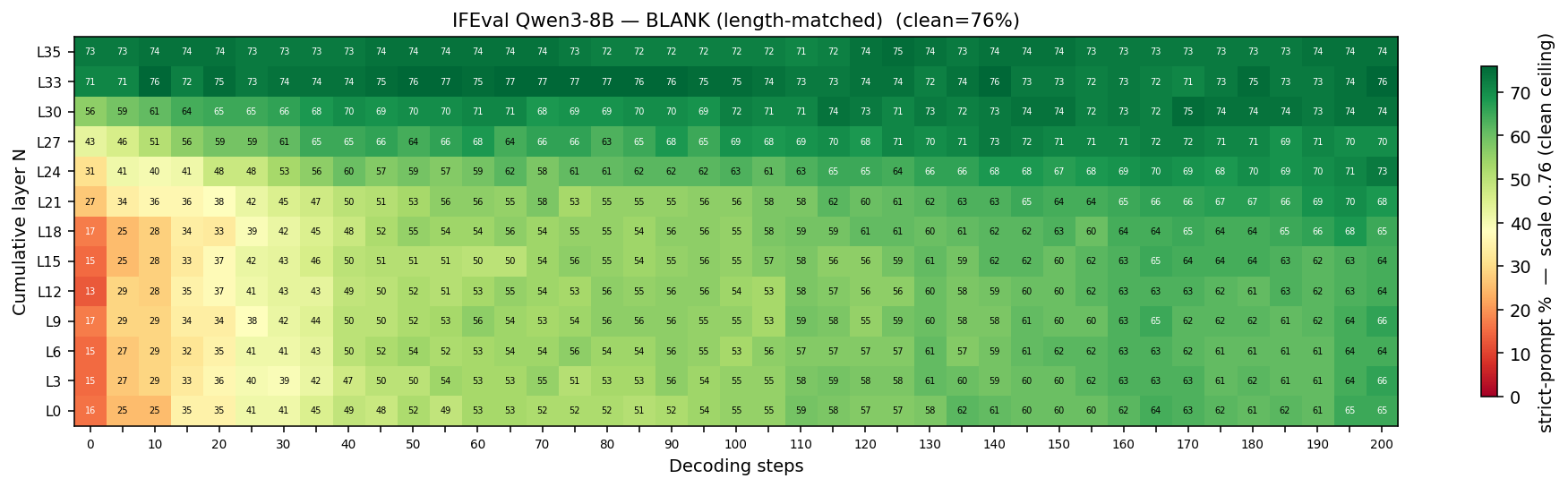}
\caption{Qwen3-8B, IFEval, \textsc{blank}}
\end{subfigure}\\[0.4em]
\begin{subfigure}[t]{0.49\linewidth}\centering
\includegraphics[width=\linewidth]{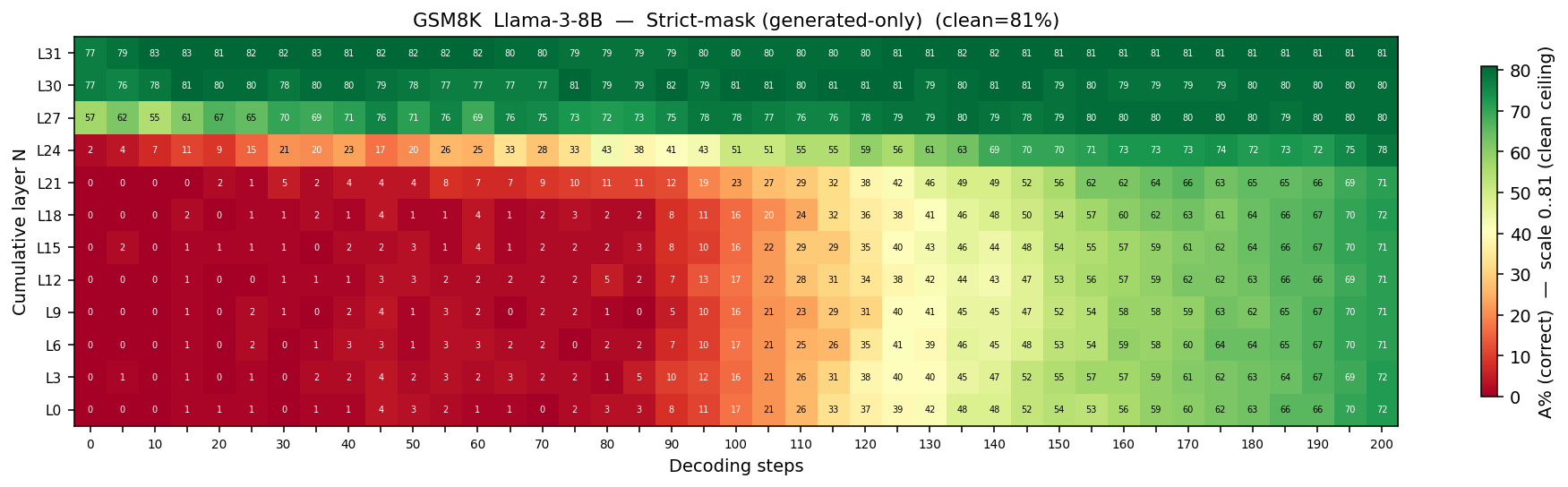}
\caption{Llama-3-8B-Instruct, GSM8K, \textsc{zero}}
\end{subfigure}\hfill
\begin{subfigure}[t]{0.49\linewidth}\centering
\includegraphics[width=\linewidth]{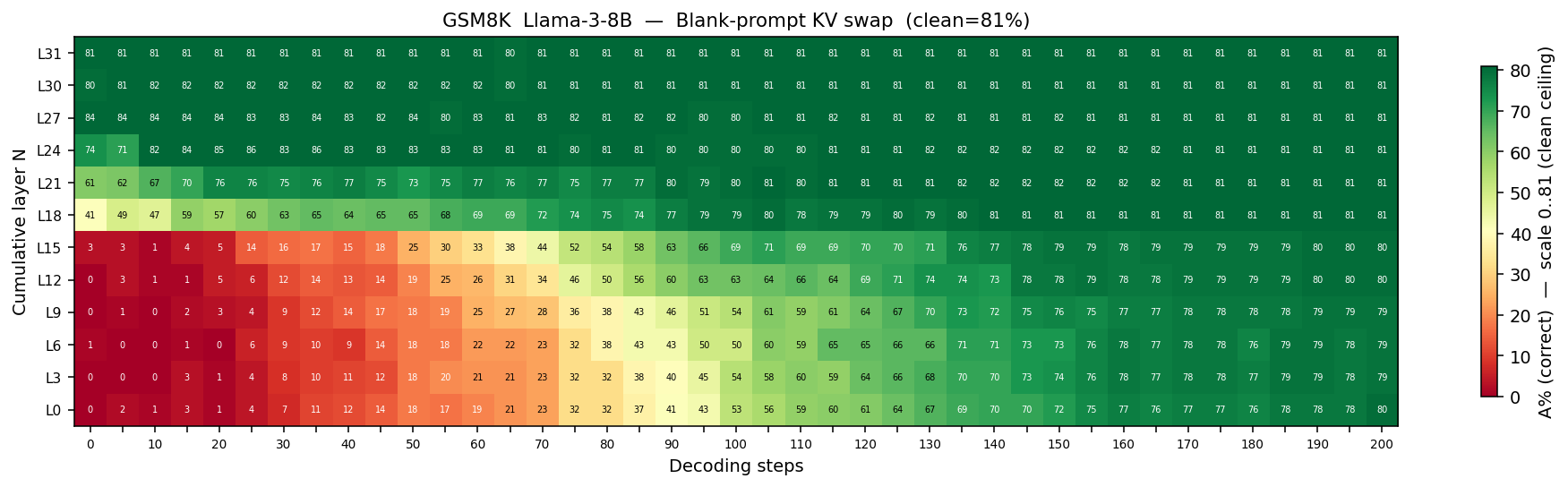}
\caption{Llama-3-8B-Instruct, GSM8K, \textsc{blank}}
\end{subfigure}\\[0.4em]
\begin{subfigure}[t]{0.49\linewidth}\centering
\includegraphics[width=\linewidth]{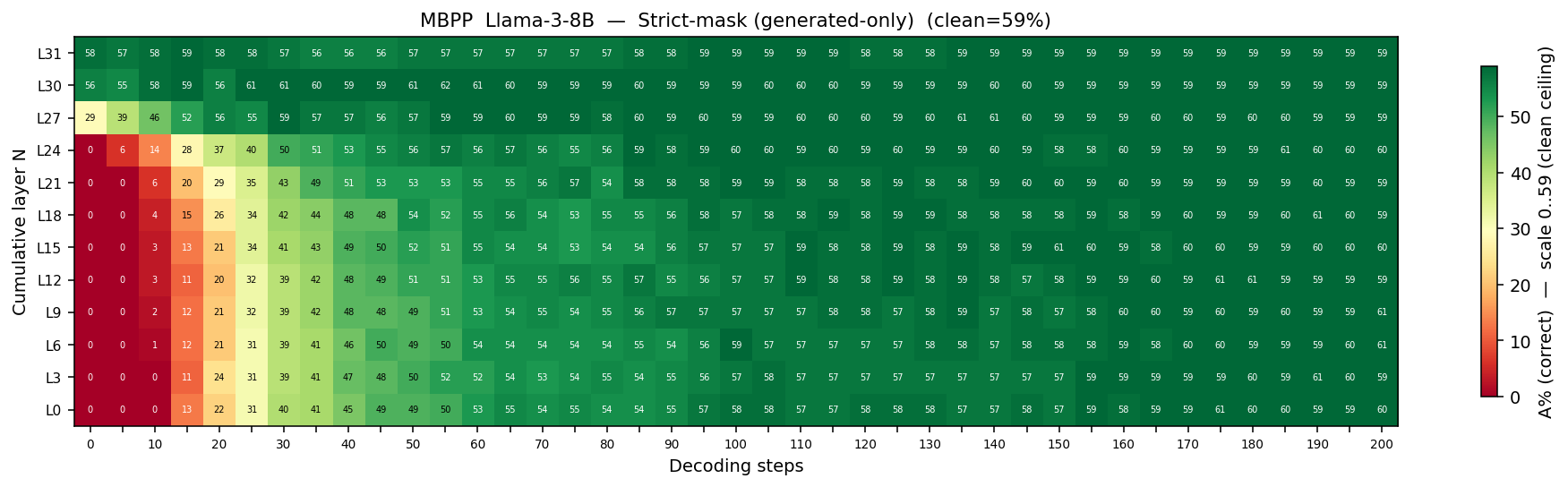}
\caption{Llama-3-8B-Instruct, MBPP, \textsc{zero}}
\end{subfigure}\hfill
\begin{subfigure}[t]{0.49\linewidth}\centering
\includegraphics[width=\linewidth]{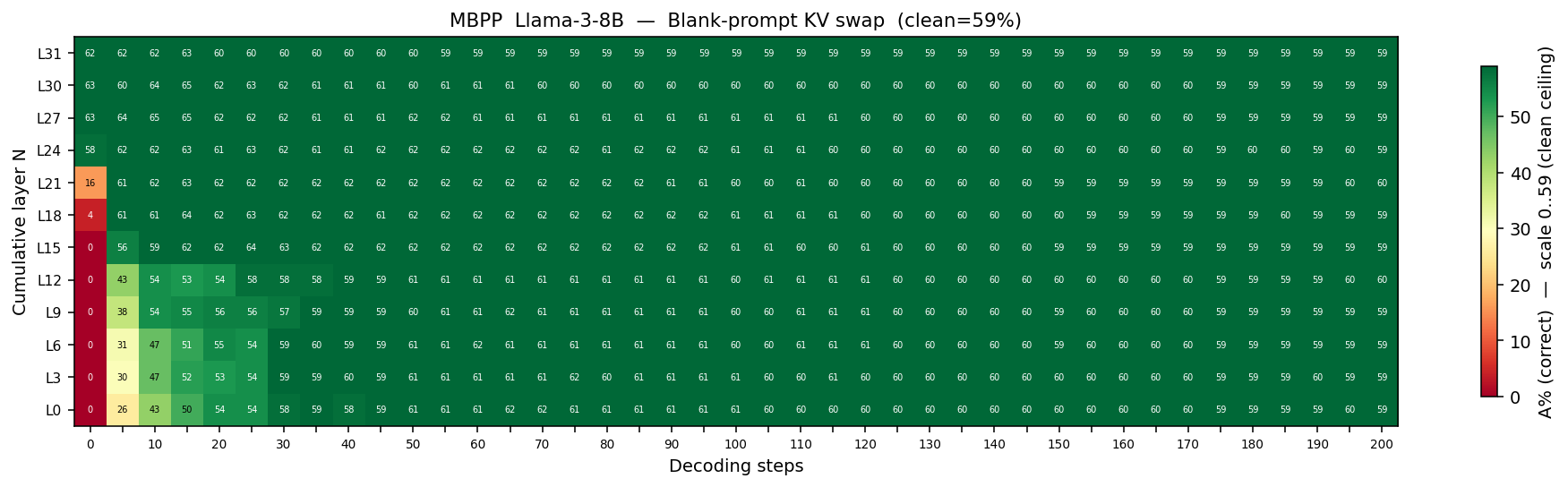}
\caption{Llama-3-8B-Instruct, MBPP, \textsc{blank}}
\end{subfigure}
\caption{Per-cell heatmaps for Gemma-3-4B-IT, Qwen3-8B, and
Llama-3-8B-Instruct on GSM8K, MBPP and IFEval. Each cell is a pass-rate
percentage at one $(L, W)$ configuration. The left column shows \textsc{zero} and the right column shows
\textsc{blank}.}
\label{fig:appendix-phase-diagrams}
\end{figure*}

\begin{figure*}[t]
\centering
\begin{subfigure}[t]{0.49\linewidth}
\centering 
\includegraphics[width=\linewidth]{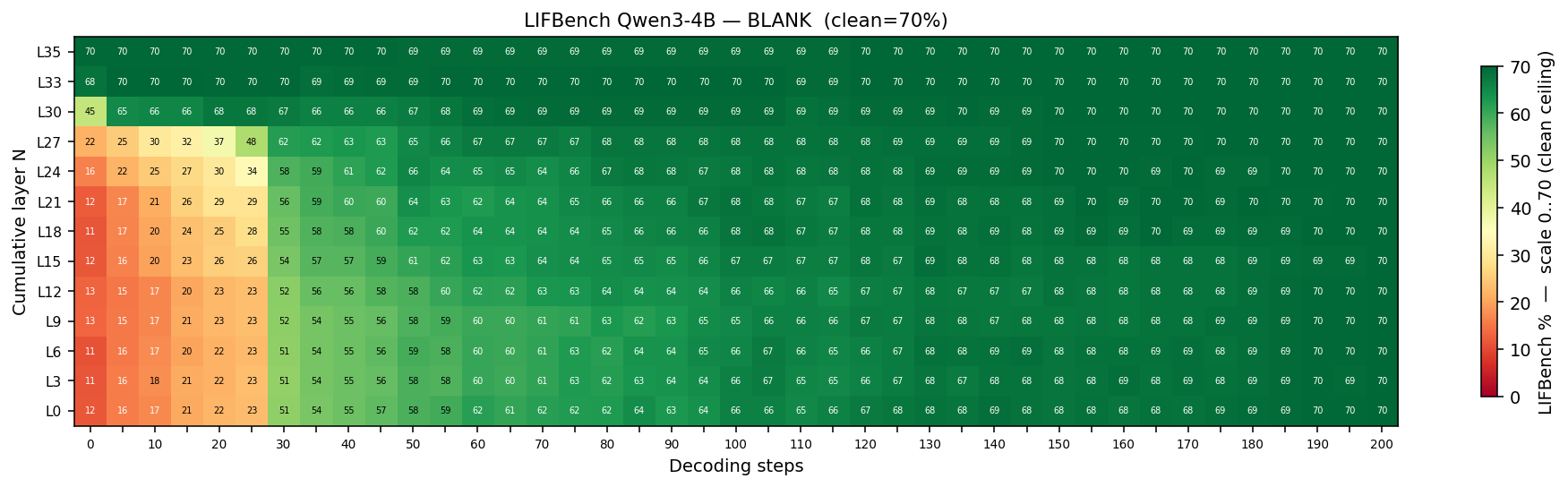}
\caption{Qwen3-4B \textsc{BLANK}}
\end{subfigure}\hfill
\begin{subfigure}[t]{0.49\linewidth}
\centering
\includegraphics[width=\linewidth]{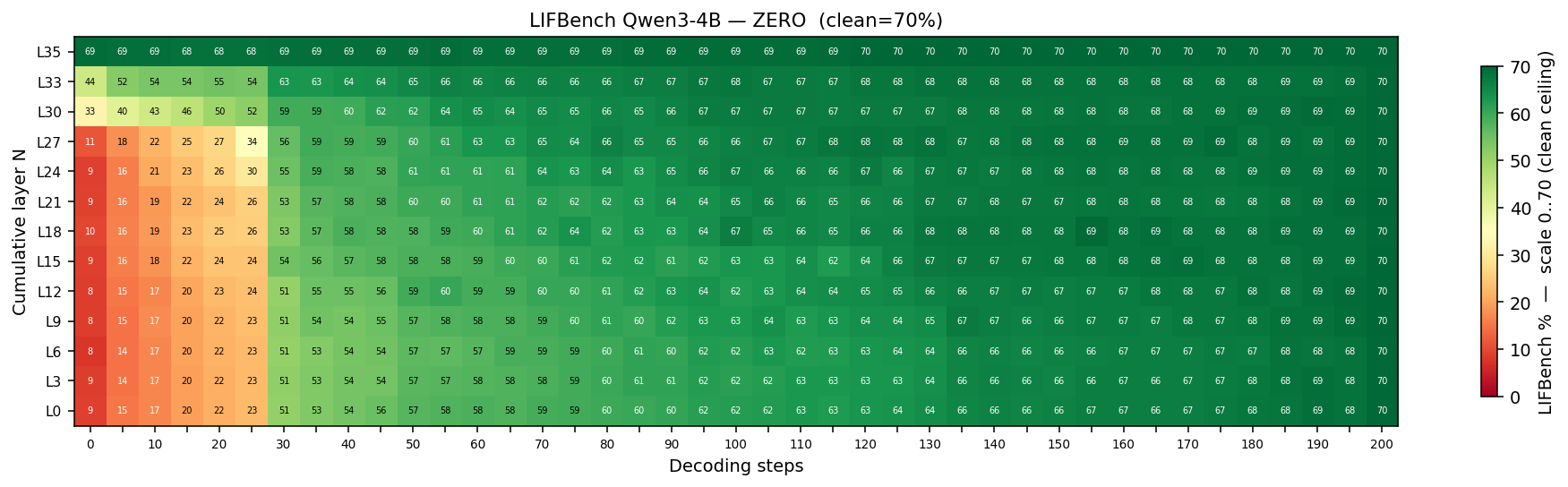}
\caption{Qwen3-4B \textsc{ZERO}}
\end{subfigure}\\[0.4em]
\begin{subfigure}[t]{0.49\linewidth}
\centering
\includegraphics[width=\linewidth]{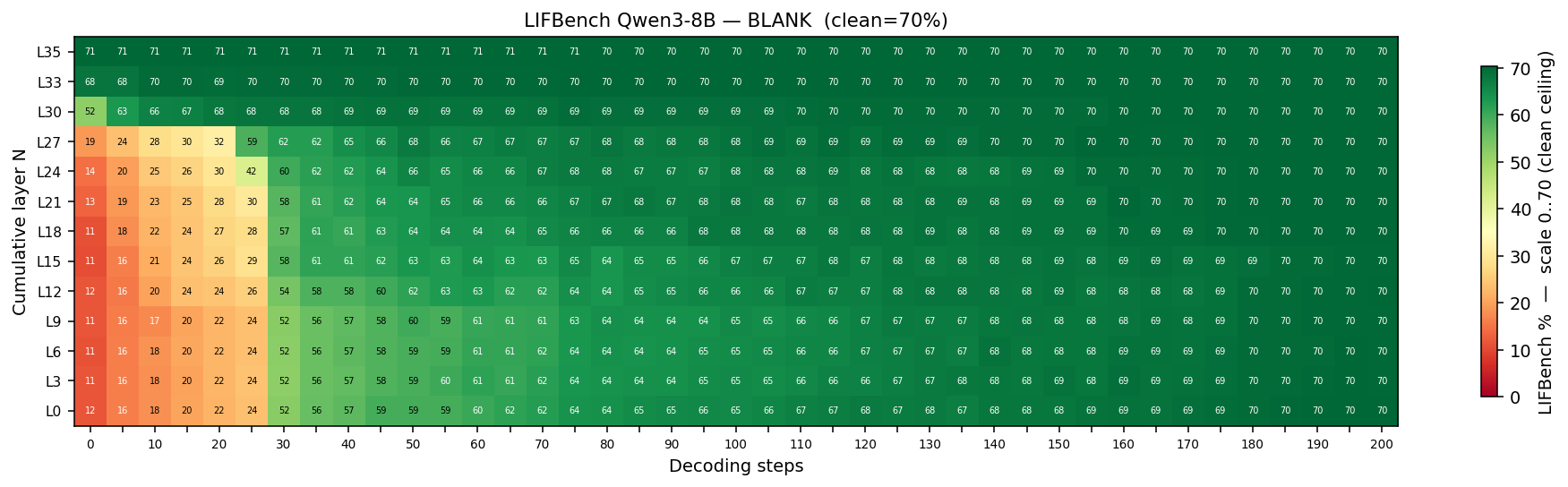}
\caption{Qwen3-8B \textsc{BLANK}}
\end{subfigure}\hfill
\begin{subfigure}[t]{0.49\linewidth}
\centering
\includegraphics[width=\linewidth]{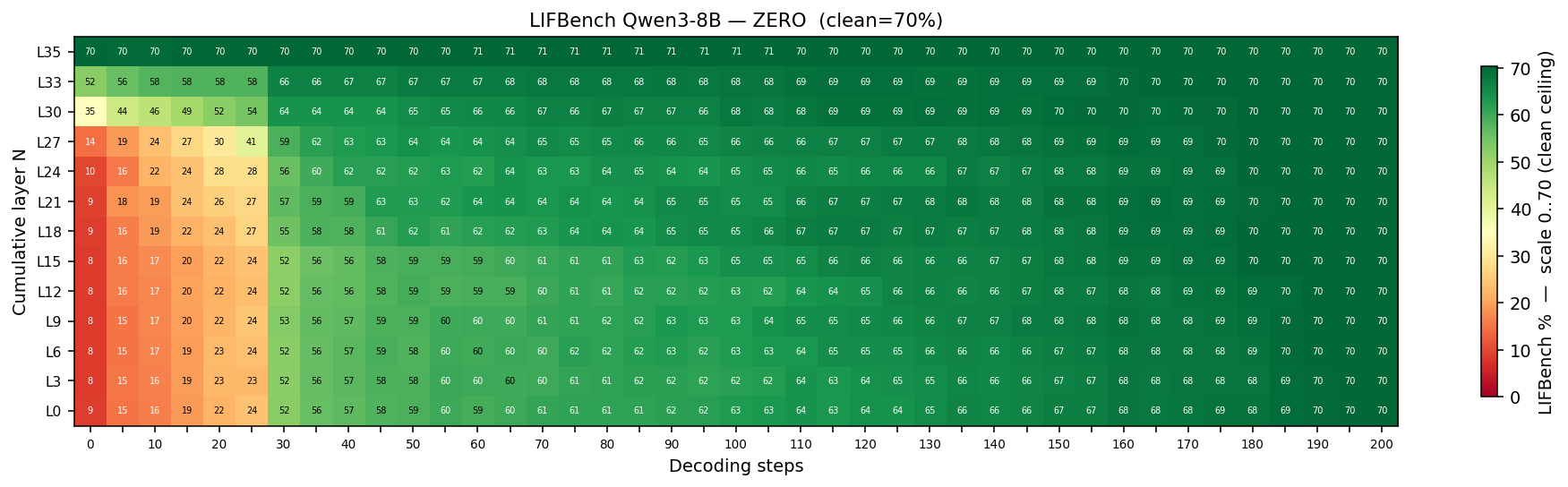}
\caption{Qwen3-8B \textsc{ZERO}}
\end{subfigure}
\caption{Qwen3-4B and Qwen3-8B heatmaps for \textsc{zero} and \textsc{blank}
on LIFBench aggregated across all lengths. }
\label{fig:long-context-lif-phase}
\end{figure*}

Fig.~\ref{fig:phase-diagrams-qwen3-4b} in the main text shows phase
diagrams for Qwen3-4B only. Fig.~\ref{fig:appendix-phase-diagrams}
provides the analogous per-cell heatmaps for Gemma-3-4B-IT, Qwen3-8B,
and Llama-3-8B-Instruct on both GSM8K and MBPP, for both \textsc{zero}
and \textsc{blank}. Across every model and task pair, the same pattern
holds. \textsc{Blank} cells stay near the clean baseline once $L$ is
sufficiently deep, while \textsc{zero} cells form a red region at
shallow $L$ that contracts only at the topmost cutoffs. The same phase structure also replicates in the long-context setting. Fig.~\ref{fig:long-context-lif-phase} shows the full $(L, W)$ grid on LIFBench (4k--16k-token prompts), where the \textsc{blank} frontier again sits left of \textsc{zero} at every cutoff.

\section{MBPP Recovery Frontier}
\label{sec:appendix-mbpp-frontier}

\begin{figure*}[t]
\centering
\includegraphics[width=0.9\linewidth]{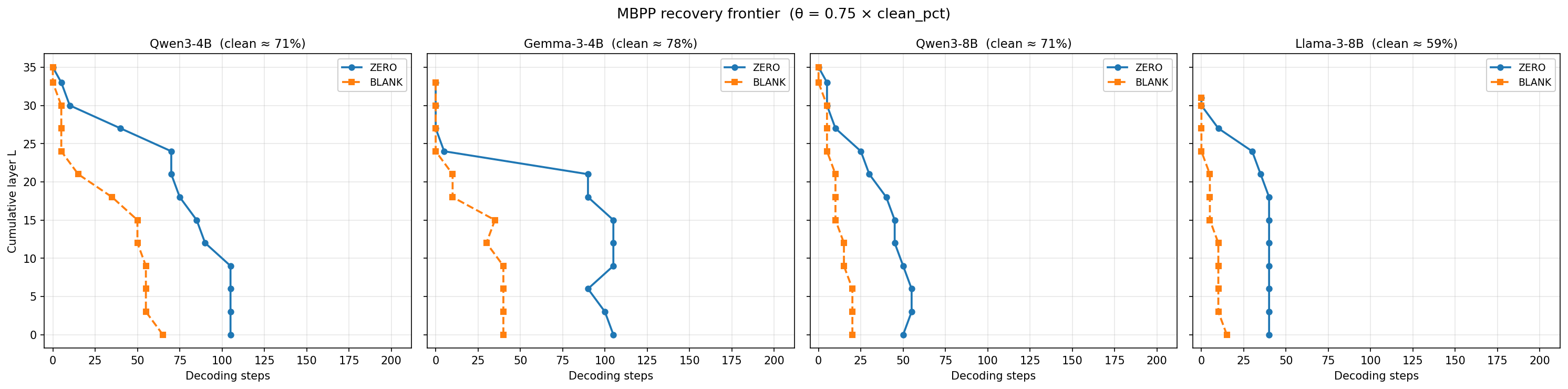}
\caption{MBPP recovery frontier $W^\star(L; \alpha\!=\!0.75)$ for
\textsc{zero} (blue) vs \textsc{blank} (orange), one panel per model.
\textsc{Blank} sits consistently to the left of \textsc{zero}, matching
the GSM8K pattern in the main text
(Fig.~\ref{fig:strict-vs-blank-frontier}).}
\label{fig:appendix-mbpp-frontier}
\end{figure*}

Fig.~\ref{fig:strict-vs-blank-frontier} in the main text shows the
GSM8K recovery frontier. Fig.~\ref{fig:appendix-mbpp-frontier} gives
the analogous $W^\star(L; \alpha\!=\!0.75)$ for \textsc{zero} vs
\textsc{blank} on MBPP across the four smaller models. The dissociation
is identical in shape: \textsc{blank} sits to the left of \textsc{zero}
in every panel, recovering with less pre-splice decoding at each $L$.

\section{Statistical Robustness of the Recovery-Frontier Gap}
\label{app:seed-variance}

\begin{figure*}[t]
  \centering
  \begin{subfigure}{0.33\textwidth}
    \includegraphics[width=\linewidth]{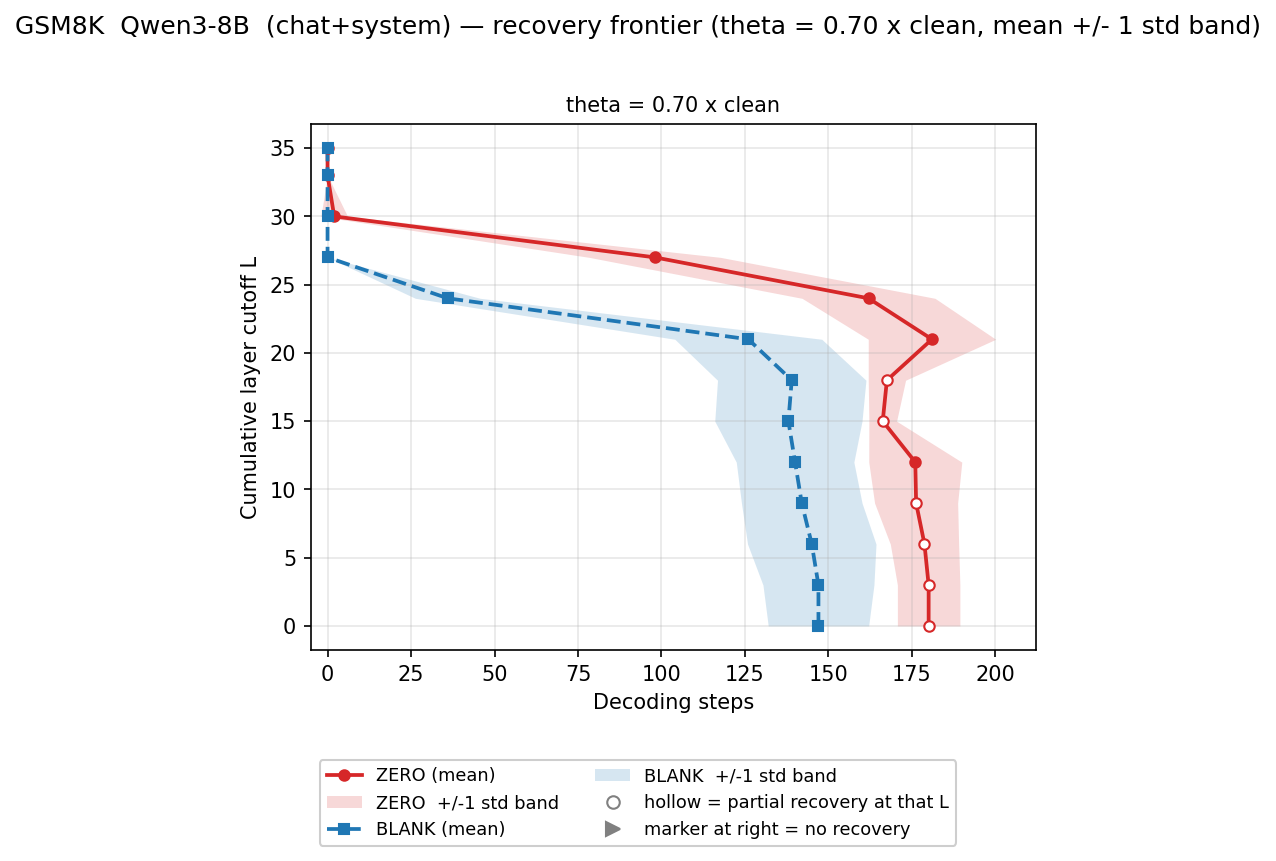}
    \caption{GSM8K, Qwen3-8B}
    \label{fig:seed-band-gsm8k-qwen8b}
  \end{subfigure}\hfill
  \begin{subfigure}{0.33\textwidth}
    \includegraphics[width=\linewidth]{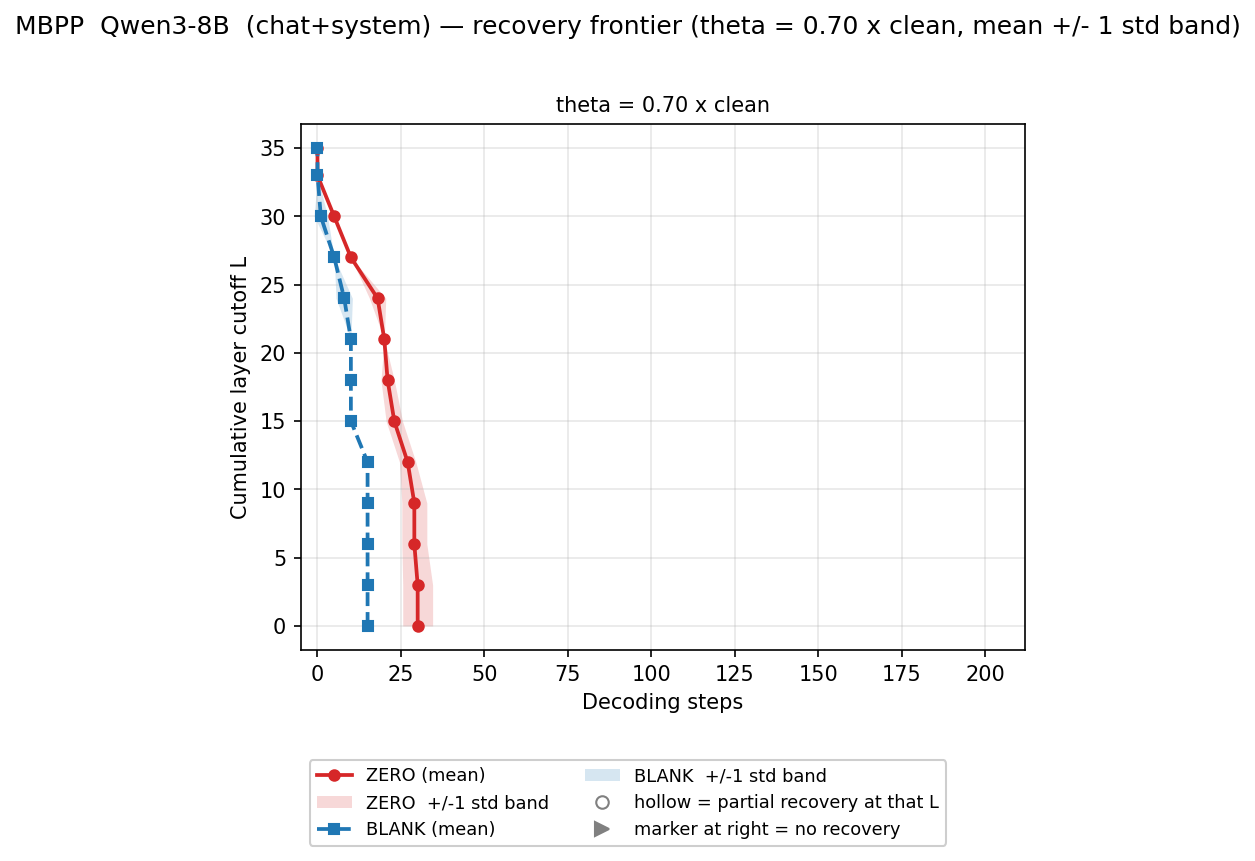}
    \caption{MBPP, Qwen3-8B}
    \label{fig:seed-band-mbpp-qwen8b}
  \end{subfigure}\hfill
  \begin{subfigure}{0.33\textwidth}
    \includegraphics[width=\linewidth]{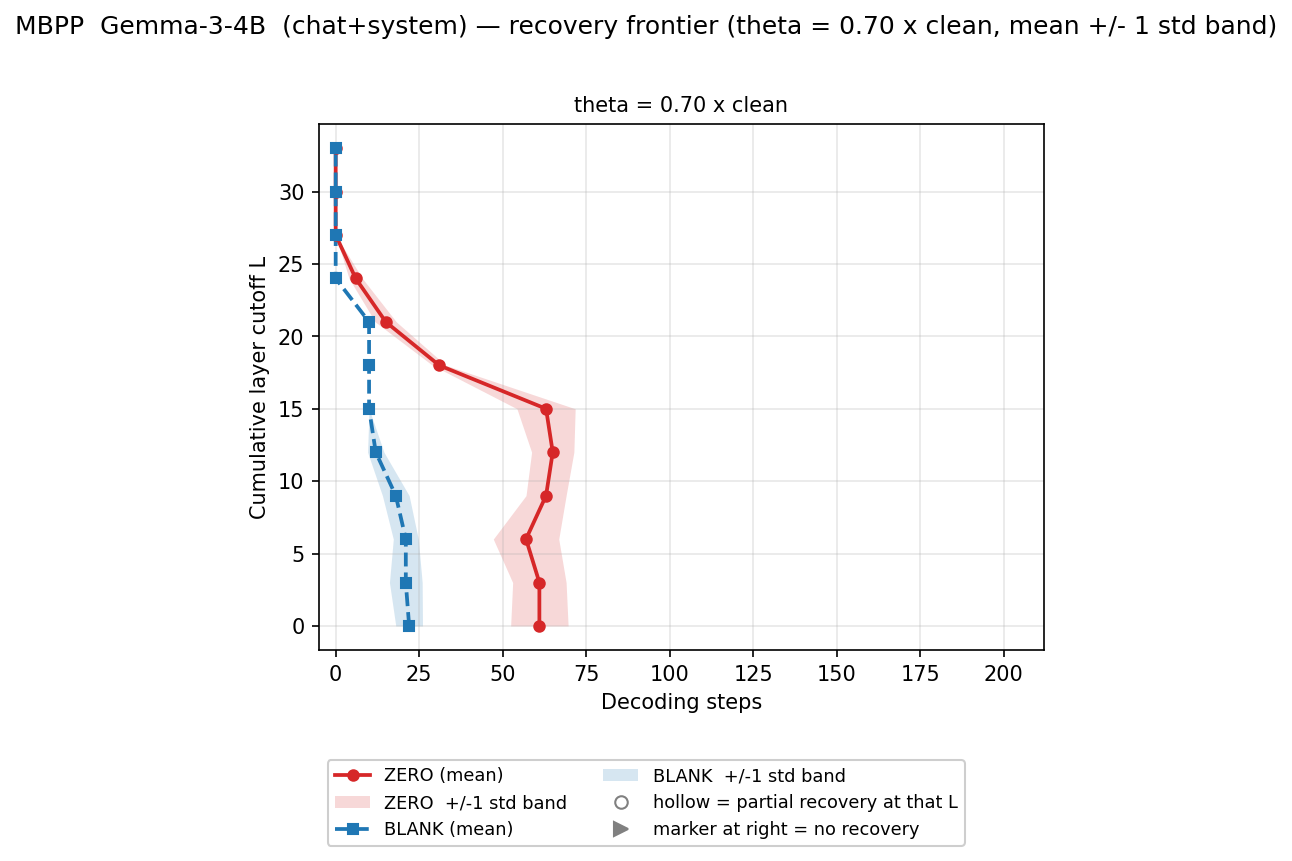}
    \caption{MBPP, Gemma-3-4B-IT}
    \label{fig:seed-band-mbpp-gemma4b}
  \end{subfigure}
  \caption{Recovery frontier $W^\star(L;\alpha)$ across five seeds under
  \textsc{Zero} (red) and \textsc{Blank} (blue); solid lines are the seed
  mean, shaded regions standard deviation. Each panel group uses
  $\alpha = 0.70$. \textsc{Blank} sits to the left of
  \textsc{Zero} at every $L$ , and the seed bands are
  narrow relative to the gap, across both tasks and model families.}
  \label{fig:seed-band-all}
\end{figure*}

To confirm that the recovery-frontier separation between \textsc{Zero} and
\textsc{Blank} is not an artifact of a single 100-problem subsample, we
recompute the frontier under five independently seeded subsamples (seeds 1
to 5) and report the seed mean with standard-deviation band, at the
recovery threshold $\theta = 0.70 \cdot p_{\text{clean}}$.
Figure~\ref{fig:seed-band-all} shows the result for Qwen3-8B on GSM8K
(\subref{fig:seed-band-gsm8k-qwen8b}) and for Qwen3-8B
(\subref{fig:seed-band-mbpp-qwen8b}) and Gemma-3-4B-IT
(\subref{fig:seed-band-mbpp-gemma4b}) on MBPP. The horizontal axis is the
splice onset $W$ (labelled warmup $T$ in the figure).

The \textsc{Blank} frontier (blue) sits to the left of the \textsc{Zero}
frontier (red) at every layer cutoff $L$ in all three panels, reproducing the
form-versus-content dissociation of (R1) and (R2) under repeated sampling.
The seed bands are narrow relative to this separation and do not overlap
across the intermediate cutoffs where the gap is largest, so the ordering is
well outside seed noise. On GSM8K the gap is wide, with \textsc{Blank}
reaching the threshold at substantially smaller $W$ than \textsc{Zero}
throughout the shallow-to-mid range, and \textsc{Zero} recovering at only a
subset of seeds at several of these cutoffs (hollow markers). On MBPP the
gap is smaller and both frontiers recover within a short warmup window, but
the \textsc{Blank}-left-of-\textsc{Zero} ordering is preserved on both
Qwen3-8B and Gemma-3-4B-IT, showing the pattern is not specific to one task
or model family. 

We conclude that the recovery-frontier gap between \textsc{Zero} and
\textsc{Blank} is a stable property across seeds, not a consequence of a
particular 100-problem draw.

\section{Clean Baselines}
\label{sec:appendix-clean}

Table~\ref{tab:clean-baselines} reports the unintervened (\emph{clean})
pass rate $p_{\mathrm{clean}}$ used as the reference in the recovery
frontier $W^\star(L; \alpha) = \min\{W : \mathrm{pass}(L, W) \geq
\alpha\,p_{\mathrm{clean}}\}$. Each entry is the pass rate of the
unmodified model on the full benchmark sample, computed once per model and task pair and shared
across all interventions.

\begin{table}[h]
\small
\centering
\setlength{\tabcolsep}{6pt}
\begin{tabular}{@{}l c c c@{}}
\toprule
model & GSM8K & MBPP & HumanEval \\
\midrule
Qwen3-4B              & 90.0 & 71.0 & 79.0 \\
Gemma-3-4B-IT         & 87.0 & 78.0 & 71.0 \\
Qwen3-8B              & 95.0 & 71.0 & 77.0 \\
Llama-3-8B-Instruct   & 81.0 & 59.0 & 59.0 \\
\bottomrule
\end{tabular}
\caption{Clean baselines $p_{\mathrm{clean}}$ (\%, exact match for GSM8K,
test-suite pass for MBPP and HumanEval) for the four models evaluated.}
\label{tab:clean-baselines}
\end{table}

\section{Splice Intervention Pseudocode}
\label{sec:appendix-pseudocode}

Algorithm~\ref{alg:splice} restates the splice intervention as a
modified greedy decoding loop. The loop differs from a standard greedy
decode only in line~\ref{alg:splice-replace}, which performs a single
in-place replacement of the prompt-span entries at the patched layers
once decoding has reached step $W+1$. The patched cache is then used for
every subsequent decoding step until generation terminates.

\begin{algorithm}[h]
\caption{Splice-intervention greedy decode}
\label{alg:splice}
\begin{algorithmic}[1]
\Require model $M$ with $N$ layers, prefilled cache $(K, V)$ with
prompt-span length $T_p$, donor cache $(\tilde{K}, \tilde{V})$,
layer cutoff $L$, splice onset $W$, max new tokens $T_{\max}$
\State $\mathcal{P} \gets \{\ell : \ell \geq L\}$ \Comment{patched layers}
\State $y_0 \gets \arg\max_v \, M(K, V)_v$
\State $\mathit{tokens} \gets [y_0]$ \Comment{t = 0}
\For{$t = 1$ to $T_{\max} - 1$}
    \If{$t= W + 1$} \label{alg:splice-replace}
        \For{$\ell \in \mathcal{P}$}
            \State $K^{(\ell)}_{0:T_p} \gets \tilde{K}^{(\ell)}_{0:T_p}$
            \State $V^{(\ell)}_{0:T_p} \gets \tilde{V}^{(\ell)}_{0:T_p}$
        \EndFor
    \EndIf
    \State $y_t \gets \arg\max_v \, M(\mathit{tokens}[-1], (K, V))$
    \If{$y_t = \mathrm{eos}$} \textbf{break} \EndIf
    \State $\mathit{tokens}.\mathrm{append}(y_t)$
\EndFor
\State \Return $\mathit{tokens}$
\end{algorithmic}
\end{algorithm}

\section{Example Generations}
\label{sec:appendix-examples}

To complement the aggregate pass-rate results,
Tables~\ref{tab:examples-mbpp} and \ref{tab:examples-gsm8k} report
side-by-side greedy completions from Qwen3-4B on representative MBPP
and GSM8K problems at $(L=27, W=0)$, the regime where the
\textsc{zero} vs \textsc{blank} gap is largest while \textsc{blank} is
still recoverable. Each row shows the output under the unintervened
decode, the \textsc{blank} intervention, and the \textsc{zero}
intervention. Visible characters are verbatim from the decoded output,
problem statements are paraphrased to a single line, and ``$\ldots$''
marks elided text. In \textsc{zero} cells this is further token-loop
repetition, and in \textsc{clean} or \textsc{blank} cells this is
surrounding prose.

\begin{table*}[h]
\small
\centering
\setlength{\tabcolsep}{4pt}
\begin{tabular}{@{}p{0.18\linewidth} p{0.26\linewidth} p{0.26\linewidth} p{0.26\linewidth}@{}}
\toprule
problem & \textsc{clean} & \textsc{blank} & \textsc{zero} \\
\midrule
{\small\texttt{string\_to\_list}: split a string on whitespace}
& \texttt{def string\_to\_list(s):\newline\hspace*{2em}return s.split()}
& \texttt{def string\_to\_list(s):\newline\hspace*{2em}return s.split()}
& \texttt{def armour\_arm\_arm\_arm\_arm\_arm\_$\ldots$} \\
\midrule
{\small\texttt{find\_star\_num}: $n$th centered hexagonal number}
& \texttt{def find\_star\_num(n):\newline\hspace*{2em}\# The formula for the n'th\newline\hspace*{2em}\# star number is 6n(n-1) + 1\newline\hspace*{2em}return 6 * n * (n - 1) + 1}
& \texttt{def a(n):\newline\hspace*{2em}return 6 * n * n - 6 * n + 1}
& \texttt{def get\_get\_get\_get\_get\_get\_$\ldots$} \\
\bottomrule
\end{tabular}
\caption{Greedy completions from Qwen3-4B on two MBPP problems under
the unintervened (\textsc{clean}) decode, the \textsc{blank}
intervention, and the \textsc{zero} intervention, all at $L=27, W=0$.
Under \textsc{blank} the model emits well-formed, mostly correct
Python. For \texttt{string\_to\_list} the completion is identical to
the unintervened output, and for \texttt{find\_star\_num} it produces
an equivalent closed form ($6n^2{-}6n{+}1$ vs.\ $6n(n{-}1){+}1$) but
under a renamed entry point (\texttt{a}), which would fail a strict
test harness despite the formula being correct. Under \textsc{zero}
the decoder collapses into a single-token loop and produces no usable
solution.}
\label{tab:examples-mbpp}
\end{table*}

\begin{table*}[h]
\small
\centering
\setlength{\tabcolsep}{4pt}
\begin{tabular}{@{}p{0.18\linewidth} p{0.26\linewidth} p{0.26\linewidth} p{0.26\linewidth}@{}}
\toprule
problem & \textsc{clean} & \textsc{blank} & \textsc{zero} \\
\midrule
{\small Dog grooming costs \$100, 30\% discount, price after?}
& \texttt{$\ldots$discount amount is \$100 $\times$ 0.30 = \$30. Subtracting$\ldots$ \$70. The answer is 70.} 
& \texttt{$\ldots$10\% of \$100 is \$10. So the discounted price is \$100 - \$10 = \$90. The answer is 90.} 
& \texttt{The original cost of the dog dog dog dog dog dog dog dog dog dog dog dog$\ldots$}  \\
\midrule
{\small Lily has 50 friends, Amy has 20 more, total?}
& \texttt{$\ldots$Amy made $50+20=70$ friends. Together $70+50=120$ friends.} 
& \texttt{$\ldots$Ava has 20 more friends than Lia$\ldots$ $50+20=70\ldots$} (truncated)
& \texttt{We know that and and and and and and and and and and and and$\ldots$}  \\
\bottomrule
\end{tabular}
\caption{Greedy completions from Qwen3-4B on two GSM8K problems under
the unintervened (\textsc{clean}) decode, the \textsc{blank}
intervention, and the \textsc{zero} intervention, all at $L=27, W=0$.
Reference (gold) answers are 70 and 120. Under \textsc{clean} the
model follows a coherent arithmetic chain and reaches the gold answer
in both cases. Under \textsc{blank} the decoder remains arithmetic and
roughly on-topic but introduces small content errors, such as
misreading the discount rate in the first problem and renaming the
protagonists in the second, resulting in incorrect final answers.
Under \textsc{zero} the model emits a degenerate single-token loop and
never produces a numeric answer.}
\label{tab:examples-gsm8k}
\end{table*}

\section{HumanEval Replication}
\label{sec:appendix-humaneval}

\begin{figure*}[p]
\centering
\includegraphics[width=0.9\linewidth]{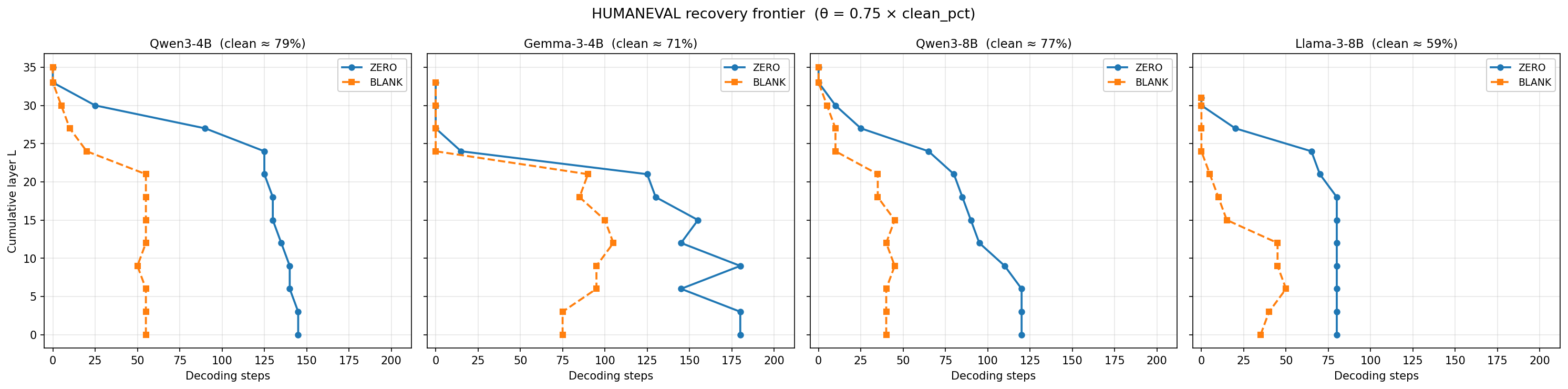}
\caption{HumanEval recovery frontier $W^\star(L; \alpha\!=\!0.75)$ for
\textsc{zero} (blue) and \textsc{blank} (orange), one panel per model.
\textsc{Blank} sits consistently to the left of \textsc{zero} across
every model.}
\label{fig:appendix-humaneval-frontier}
\end{figure*}

\begin{figure*}[p]
\centering
\begin{subfigure}[t]{0.49\linewidth}\centering
\includegraphics[width=\linewidth]{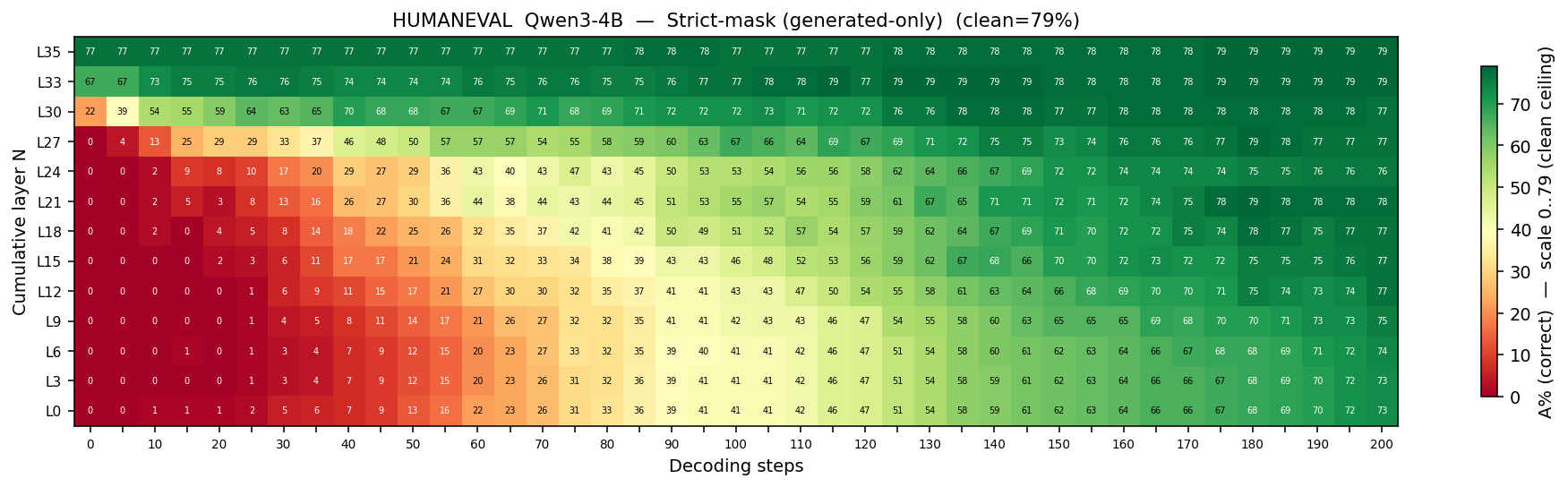}
\caption{Qwen3-4B, HumanEval, \textsc{zero}}
\end{subfigure}\hfill
\begin{subfigure}[t]{0.49\linewidth}\centering
\includegraphics[width=\linewidth]{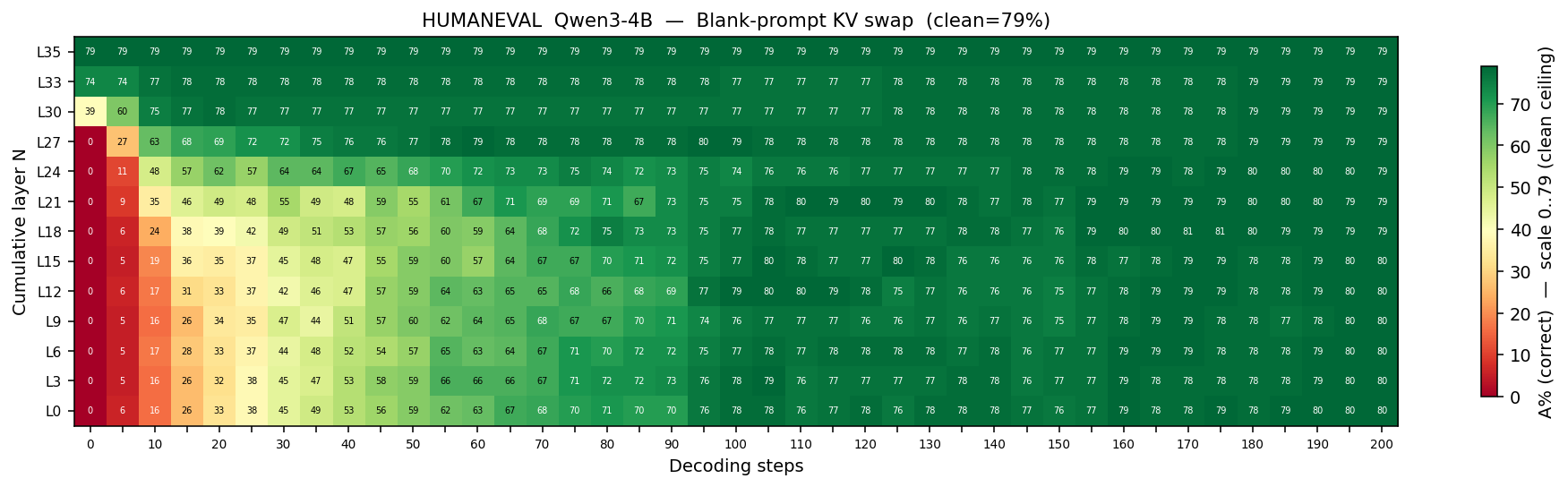}
\caption{Qwen3-4B, HumanEval, \textsc{blank}}
\end{subfigure}\\[0.4em]
\begin{subfigure}[t]{0.49\linewidth}\centering
\includegraphics[width=\linewidth]{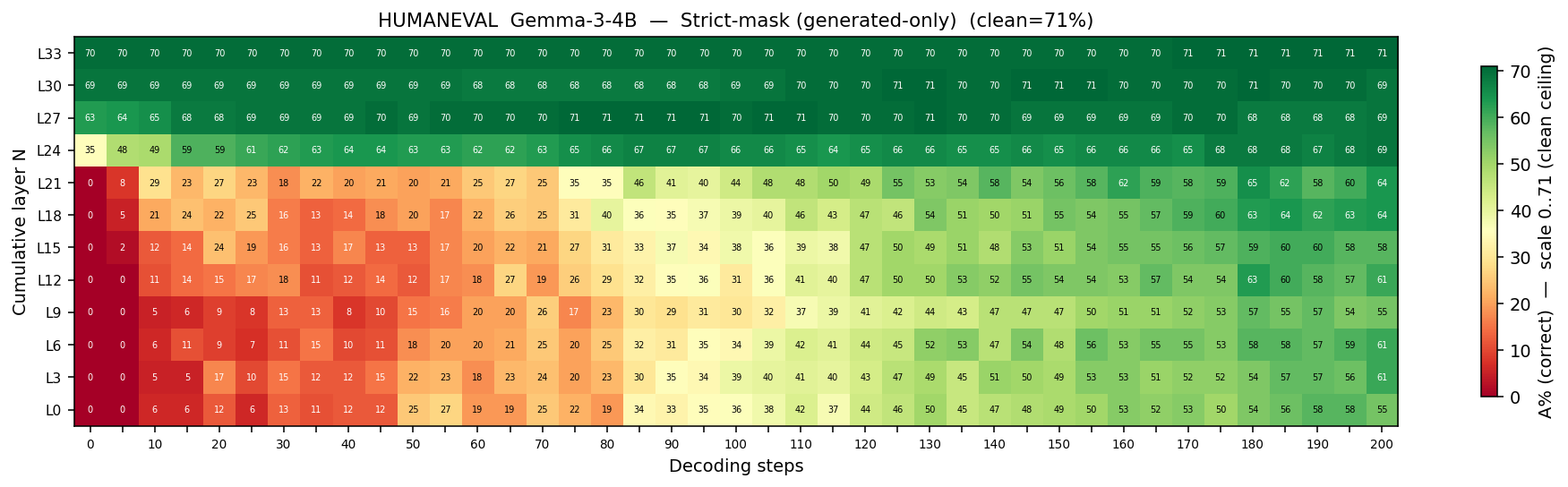}
\caption{Gemma-3-4B-IT, HumanEval, \textsc{zero}}
\end{subfigure}\hfill
\begin{subfigure}[t]{0.49\linewidth}\centering
\includegraphics[width=\linewidth]{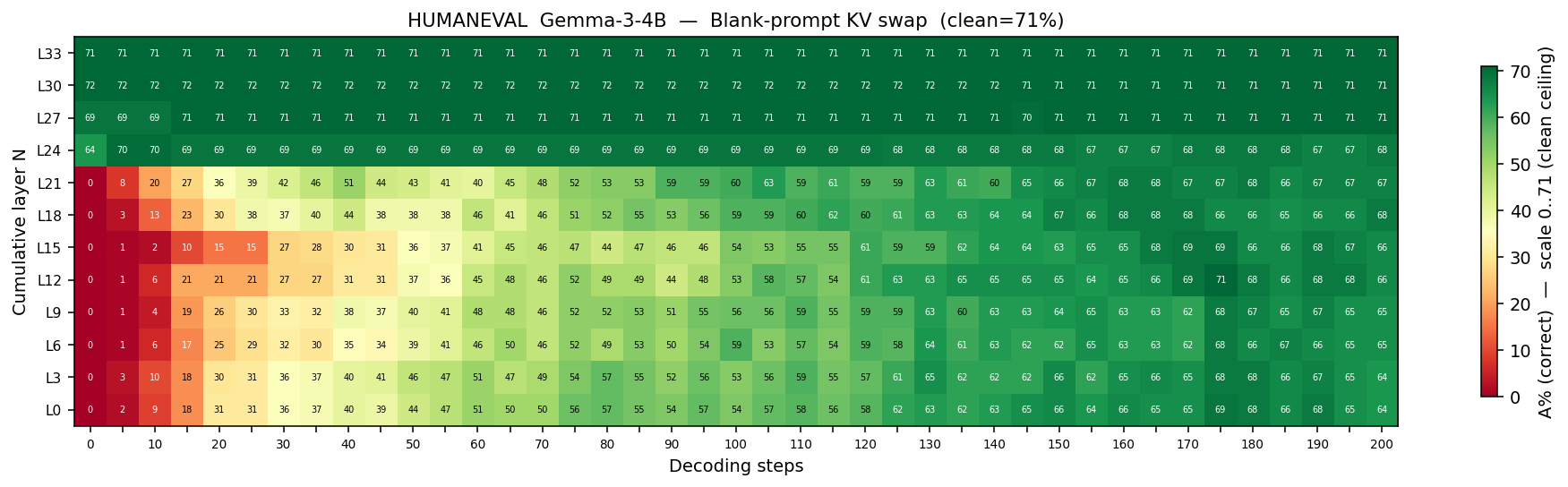}
\caption{Gemma-3-4B-IT, HumanEval, \textsc{blank}}
\end{subfigure}\\[0.4em]
\begin{subfigure}[t]{0.49\linewidth}\centering
\includegraphics[width=\linewidth]{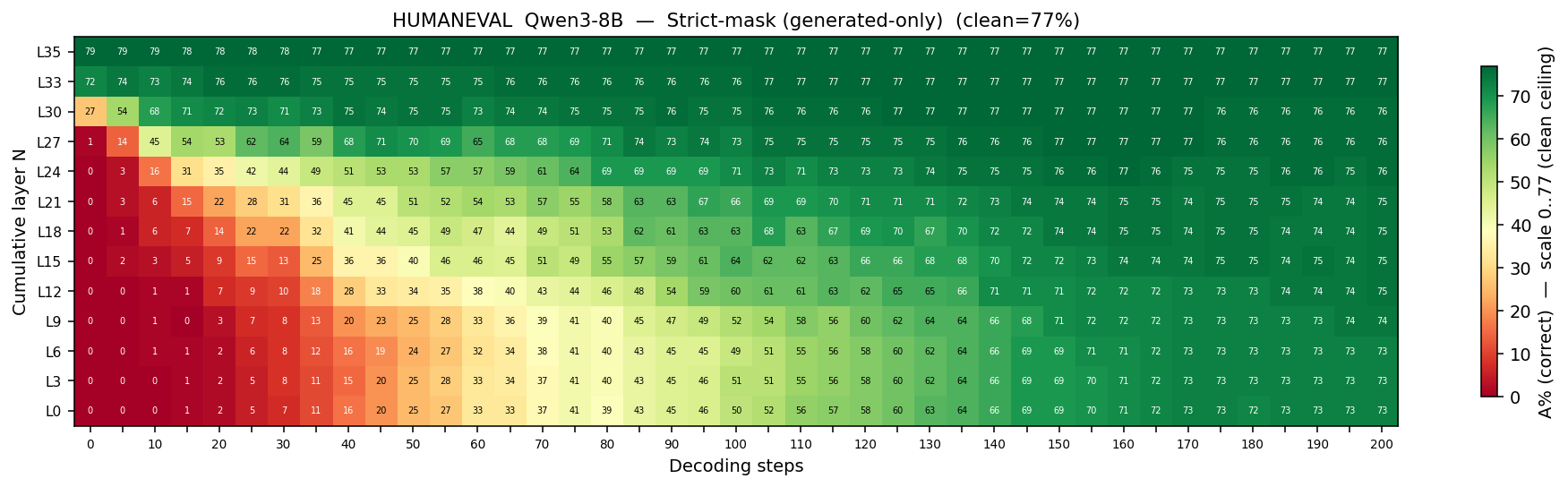}
\caption{Qwen3-8B, HumanEval, \textsc{zero}}
\end{subfigure}\hfill
\begin{subfigure}[t]{0.49\linewidth}\centering
\includegraphics[width=\linewidth]{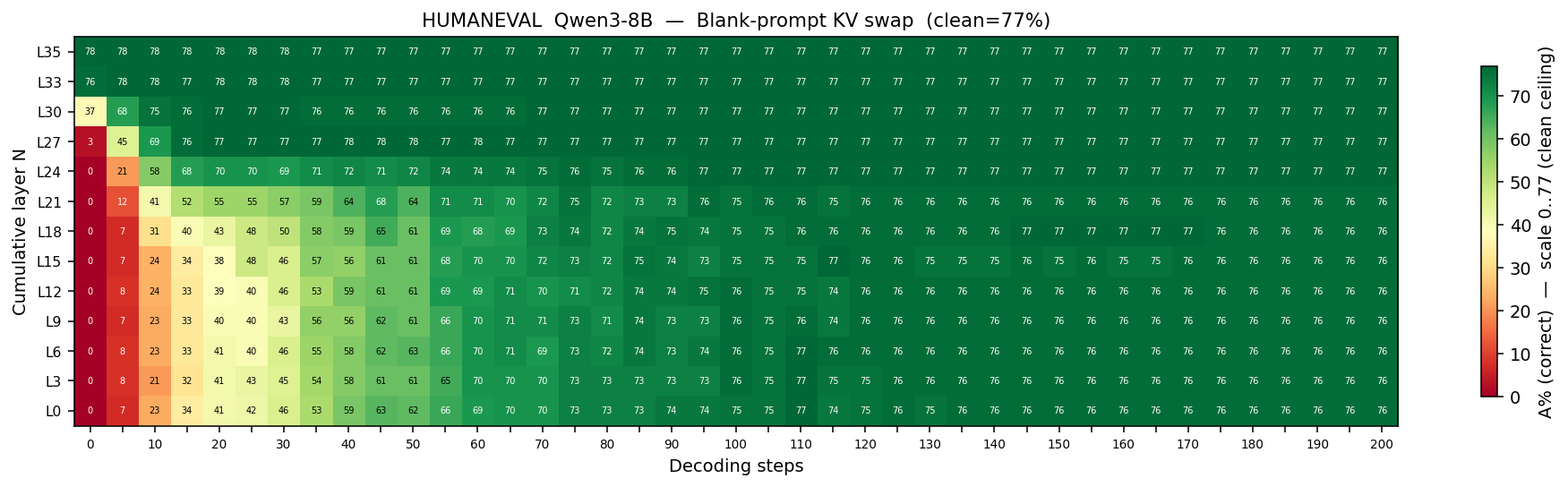}
\caption{Qwen3-8B, HumanEval, \textsc{blank}}
\end{subfigure}\\[0.4em]
\begin{subfigure}[t]{0.49\linewidth}\centering
\includegraphics[width=\linewidth]{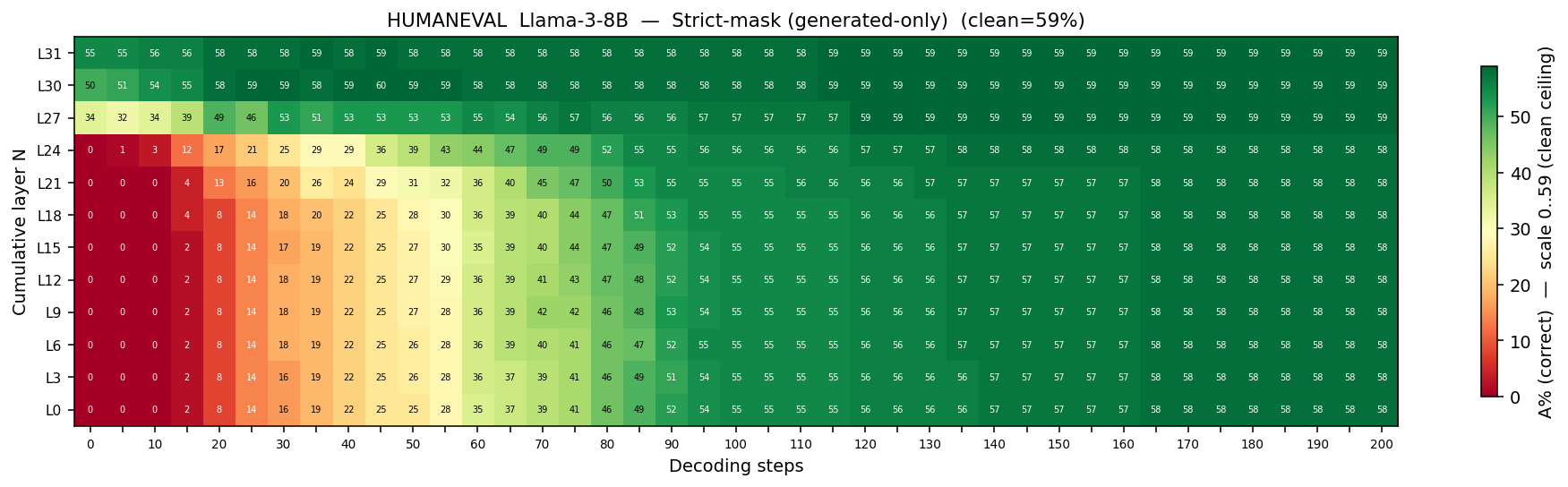}
\caption{Llama-3-8B-Instruct, HumanEval, \textsc{zero}}
\end{subfigure}\hfill
\begin{subfigure}[t]{0.49\linewidth}\centering
\includegraphics[width=\linewidth]{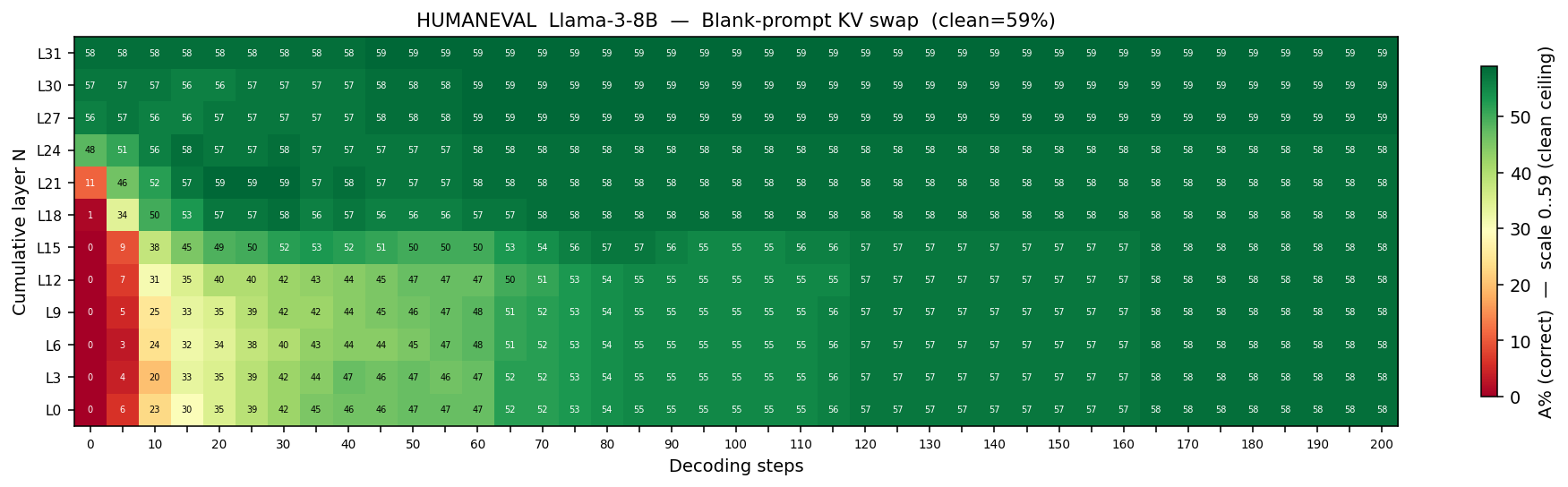}
\caption{Llama-3-8B-Instruct, HumanEval, \textsc{blank}}
\end{subfigure}
\caption{Per-cell heatmaps on HumanEval for the four models, one row
per model. The left column shows \textsc{zero} and the right column
shows \textsc{blank}. Each cell is a pass-rate percentage at one
$(L, W)$ configuration. Rows within a panel are cumulative layer
cutoffs, columns are pre-splice decoding steps $W$. \textsc{Blank}
stays near the clean baseline at deep $L$, while \textsc{zero}
produces a large red region at shallow $L$ that only contracts at the
topmost cutoffs.}
\label{fig:appendix-humaneval-heatmaps}
\end{figure*}

The form-vs-content dissociation also replicates on
HumanEval~\citep{chen2021humaneval}. Fig.~\ref{fig:appendix-humaneval-frontier}
shows the recovery frontier $W^\star(L; \alpha\!=\!0.75)$ for
\textsc{zero} and \textsc{blank} on all four models, and
Fig.~\ref{fig:appendix-humaneval-heatmaps} gives the per-cell phase
diagrams. Across every model, \textsc{blank} sits to the left of
\textsc{zero} and reaches near-baseline at deep $L$ while \textsc{zero}
requires substantially more pre-splice decoding to recover.

\section{Per-Model Heatmaps on the Algorithmic-Donor Benchmark}
\label{sec:appendix-algo-heatmaps}

\begin{figure*}[p]
\centering
\includegraphics[width=\linewidth]{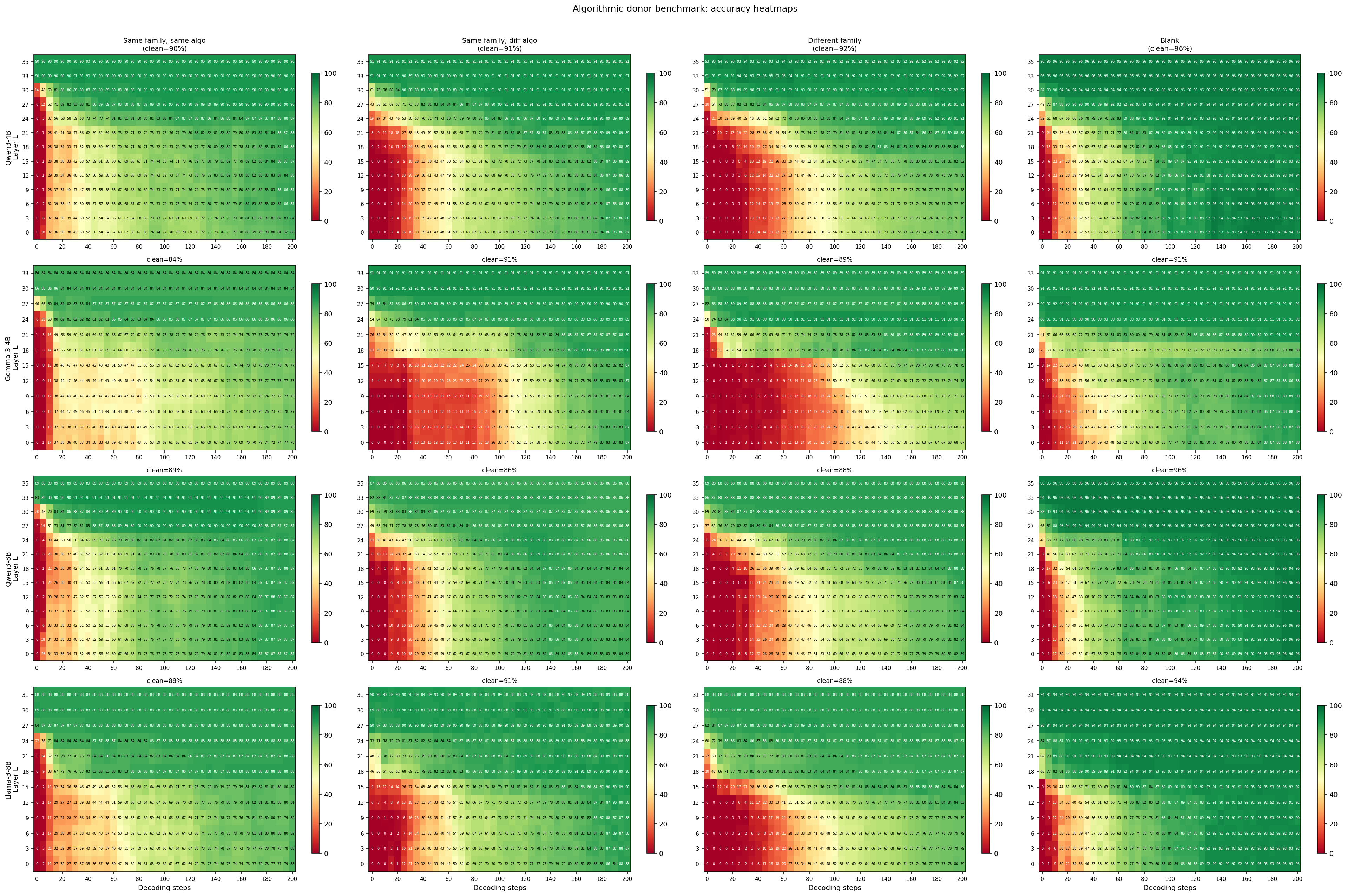}
\caption{Unified per-cell heatmaps on the algorithmic-donor benchmark.
Rows are models (Qwen3-4B, Gemma-3-4B-IT, Qwen3-8B,
Llama-3-8B-Instruct), and columns group the four donor caches
(\textsc{blank}, \textsc{same-algo}, \textsc{diff-algo},
\textsc{diff-family}) under test-pass scoring and an algo-aware score
that also requires the emitted solution to use the prescribed
algorithm class.}
\label{fig:appendix-algo-unified}
\end{figure*}

Fig.~\ref{fig:appendix-algo-unified} gives a unified cell-level view
of the algorithmic-donor benchmark across the four models and four
donor caches, under both test-pass scoring and an algo-aware score
that additionally checks whether the emitted solution uses the
prescribed algorithm class. Reading across columns reproduces the
noise ladder of (R3). Damage grows from \textsc{blank} and
\textsc{same-algo} through \textsc{diff-algo} to \textsc{diff-family}.
The two scoring rules diverge most under \textsc{diff-algo}, where the
model passes the tests but commits to the donor's algorithm class, so
the algo-aware column develops a large red region that persists to
deeper $L$. 

\end{document}